# Irrelevant and Independent Natural Extension for Sets of Desirable Gambles

**Gert de Cooman**　　　　　　　　　　　　　　　　　　　　　　GERT.DECOOMAN@UGENT.BE
*Ghent University, SYSTeMS Research Group*
*Technologiepark 914, 9052 Zwijnaarde, Belgium*

**Enrique Miranda**　　　　　　　　　　　　　　　　　　　　　MIRANDAENRIQUE@UNIOVI.ES
*University of Oviedo, Dept. of Statistics and O.R.*
*C-Calvo Sotelo, s/n, Oviedo 33007, Spain*

## Abstract

The results in this paper add useful tools to the theory of sets of desirable gambles, a growing toolbox for reasoning with partial probability assessments. We investigate how to combine a number of marginal coherent sets of desirable gambles into a joint set using the properties of epistemic irrelevance and independence. We provide formulas for the smallest such joint, called their independent natural extension, and study its main properties. The independent natural extension of maximal coherent sets of desirable gambles allows us to define the strong product of sets of desirable gambles. Finally, we explore an easy way to generalise these results to also apply for the conditional versions of epistemic irrelevance and independence. Having such a set of tools that are easily implemented in computer programs is clearly beneficial to fields, like AI, with a clear interest in coherent reasoning under uncertainty using general and robust uncertainty models that require no full specification.

## 1. Introduction

In reasoning and decision making under uncertainty, there is little doubt that probabilities play the leading part. Imprecise probability models provide a well-founded extension to probabilistic reasoning, that allow us to deal with incomplete probability assessments, indecision and robustness issues.[1]

Early imprecise probability models (going back to, amongst others, Bernoulli, 1713; Boole, 1952, 1961; Koopman, 1940) centered on lower and upper probabilities for events or propositions. In later stages (see for instance Smith, 1961; Williams, 1975b and, for the clearest statement, Walley, 1991, Section 2.7), it became apparent that the language of events and *lower probabilities* is lacking in power of expression, and that a much more expressive theory can be built using random variables and *lower previsions* (or lower expectations), instead.[2] However, even though it has been quite successful, and is by now quite well developed, there are a number of problems with the lower prevision approach. Its mathematical complexity is fairly high, especially when conditioning and independence enter the picture. Also, the coherence requirements, which specify basic rules for proper inference using (conditional) lower previsions, are quite cumbersome, and rather harder to chop down into intuitive elementary building blocks than their precise-probabilistic counterparts, even though the latter turn out to be special instances of the former. Finally, as is the case with many other approaches to probability, and as we will see further on, the theory of coherent lower previsions has issues with conditioning on sets of probability zero.

---

1. To get a good idea of what the field of imprecise probabilities is about, and how it is evolving, browse through the online proceedings of the biennial ISIPTA conferences, to be found on the web site (www.sipta.org) of the Society for Imprecise Probability: Theories and Applications.
2. In contrast, for precise probability models, the expressive power of probabilities and expectations is the same: a linear prevision or expectation on the set of all (bounded) real-valued maps is uniquely determined by its restriction to events (a finitely additive probability), and vice versa.

A very attractive solution to these problems was offered by Walley (2000), in the form of sets of desirable gambles. Walley's work was inspired by earlier ideas by Smith (1961) and Williams (1975b), but previous work along these lines was also done by Seidenfeld, Schervish, and Kadane (1995). On this approach, the primitive notions are not probabilities of events, nor expectations of random variables. Rather, the starting point is the question whether a gamble, or a risky transaction, is desirable to a subject, i.e. strictly preferred to the zero transaction, or status quo. A basic belief model is then not a probability measure, nor a lower prevision, but a *set of desirable gambles*.

Let us briefly summarise here why we believe working with sets of desirable gambles as basic belief models deserves more attention in the AI community:

*Primo*, as a number of examples in the literature have shown (Couso & Moral, 2011; De Cooman & Quaeghebeur, 2012; Moral, 2005), and as we shall see further on (look for instance at Examples 1 and 2), working with and making inferences using a *set of desirable gambles* as a subject's uncertainty model is more general and more expressive. It is also arguably simpler and more elegant from a mathematical point of view, and it has a very intuitive geometrical interpretation (Quaeghebeur, 2012b).

*Secundo*, we shall see in Sections 4 and 5 that the approach to coherent marginalisation and conditioning is especially straightforward, and there are no issues with conditioning on sets of probability zero.

*Tertio*, as we will argue in Section 2.3, because of the similarity between accepting a gamble on the one hand, and accepting a proposition to be true on the other, working with sets of desirable gambles leads to an account of probabilistic inference with a very 'logical' flavour; see the work by Moral and Wilson (1995) for an early discussion of this idea.

*Quarto*, working with sets of desirable gambles encompasses and subsumes as special cases both classical (or 'precise') probabilistic inference and inference in classical propositional logic; see Sections 2 and 5.

And finally, *quinto*, as will be made clear by the discussion throughout, sets of desirable gambles are eminently suited for dealing with partial probability assessments, in situations where experts express their beliefs, preferences or behavioural dispositions using finitely many assessments that need not determine a unique probability measure. In particular, we will discuss the connection with partial preferences in Section 2.1.

Let us try and present a preliminary defense of these sweeping claims with a few examples. One particular perceived disadvantage of working with lower previsions—or with previsions and probabilities for that matter—is that conditioning a lower prevision need not lead to uniquely coherent results when the conditioning event has lower or upper probability zero; see for instance the work of Walley (1991, Section 6.4). For precise probabilities, this difficulty can be circumvented by using full conditional measures (Dubins, 1975). As we have already mentioned, in an imprecise-probabilities context, working with the more informative coherent sets of desirable gambles rather than with lower previsions provides a very elegant and intuitively appealing way out of this problem, as it has already been suggested by Walley (1991, Section 3.8.6 and Appendix F), and argued in much more detail in his later work (Walley, 2000). The connection between full conditional measures and maximal coherent sets of desirable gambles was recently explored by Couso and Moral (2011): the latter are still more general and expressive.

The work by De Cooman and Quaeghebeur (2012) has shown that working with sets of desirable gambles is especially illuminating in the context of modelling exchangeability assessments: it exposes the simple geometrical meaning of the notion of exchangeability, and leads to a simple and particularly elegant proof of a significant generalisation to de Finetti's representation theorem for exchangeable random variables (de Finetti, 1931).

Exchangeability is a structural assessment, and so is independence, quite common in the context of probabilistic graphical models, such as Bayesian (Pearl, 1985) or credal networks (Cozman, 2000). Conditioning and independence are, of course, closely related. In a recent paper (De Cooman, Miranda, & Zaffalon, 2011), we investigated the notions of epistemic independence of finite-valued variables using coherent lower previsions, thus adding to the literature where assessments of epistemic irrelevance and independence are studied for graphical models (De Cooman, Hermans, Antonucci, & Zaffalon, 2010; Destercke & De Cooman, 2008), as an alternative to the more often used notion of strong independence. The above-mentioned problems with conditioning, and the fact that the coherence requirements for conditional lower previsions are, to be honest, quite cumbersome to work with, have turned this into a quite complicated exercise. This is the reason why, in the present paper, we intend to show that looking at independence using sets of desirable gambles leads to a more elegant theory that avoids some of the complexity pitfalls of working with coherent lower previsions. In doing this, we build on the strong pioneering work on epistemic irrelevance by Moral (2005). While we focus here on the symmetrised notion of epistemic independence, much of what we do can be seen as an application and continuation of his ideas.

Our goal in this paper is to show how local models for some variables, together with independence assessments, can be combined in order to produce a joint model. This joint model can then be used to draw inferences, as is done for instance in the context of Bayesian or credal networks (Antonucci, de Campos, & Zaffalon, 2012; Cozman, 2000; Pearl, 1985). One of the core ideas of such probabilistic graphical models is to provide a representation of this joint model that is less taxing from a computational point of view.

There are three main novelties to our approach: the first is that we allow for imprecision in the local models—although precise models are a particular case; the second is that we model local probability assessments by means of sets of desirable gambles, because of the above-mentioned advantages they possess over coherent lower previsions; and the third is that we stress epistemic irrelevance and independence rather than the more common assessment strong independence, for reasons that will become clear further on—although we also discuss strong independence.

With the results in this paper we are adding useful tools to the growing toolbox for reasoning with partial probability assessments that sets of desirable gambles constitute, something already started in our work on exchangeability (De Cooman & Quaeghebeur, 2012) and the work on epistemic irrelevance and credal networks by Moral (2005). In this regard, it is also interesting to mention that algorithms for making inferences with sets of desirable gambles have been recently established (Couso & Moral, 2011; Quaeghebeur, 2012a). Having such a set of tools that are easily implemented in computer programs is clearly beneficial to a field like AI, which should surely be interested in coherent reasoning under uncertainty with general and robust uncertainty models that require no full specification. This paper constitutes a further step in that direction, and it also allows us to see more clearly which are the main difficulties faced when working with sets of desirable gambles. There remain, however, a number of important situations to be dealt with, and future lines of research are discussed in a number of places in the paper, as well as in the Conclusion.

In Section 2 we summarise relevant results in the existing theory of sets of desirable gambles. After mentioning useful notational conventions in Section 3, we recall the basic marginalisation, conditioning and extension operations for sets of desirable gambles in Sections 4 and 5. We use these to combine a number of marginal sets of desirable gambles into a joint satisfying epistemic irrelevance (Section 6), and epistemic independence (Section 7). In Section 8, we study the particular case of maximal coherent sets of desirable gambles, and derive the concept of a strong product. Section 9 deals with conditional independence assessments.

## 2. Coherent Sets of Desirable Gambles and Natural Extension

Let us begin by explaining what our basic uncertainty models, coherent sets of desirable gambles, are about (more details can be found in Augustin, Coolen, De Cooman, & Troffaes, 2012; Couso & Moral, 2011; De Cooman & Quaeghebeur, 2012; Moral, 2005; Walley, 2000).

Consider a variable $X$ taking values in some possibility space $\mathcal{X}$, which we assume in this paper to be *finite*.[3] We model information about $X$ by means of sets of desirable gambles. A *gamble* is a real-valued function on $\mathcal{X}$, and we denote the set of all gambles on $\mathcal{X}$ by $\mathcal{G}(\mathcal{X})$. It is a linear space under point-wise addition of gambles, and point-wise multiplication of gambles with real numbers. For any subset $\mathcal{A}$ of $\mathcal{G}(\mathcal{X})$, we denote by $\mathrm{posi}(\mathcal{A})$ the set of all positive linear combinations of gambles in $\mathcal{A}$:

$$\mathrm{posi}(\mathcal{A}) := \bigg\{ \sum_{k=1}^{n} \lambda_k f_k \colon f_k \in \mathcal{A},\ \lambda_k > 0,\ n > 0 \bigg\}.$$

We call $\mathcal{A}$ a *convex cone* if it is closed under positive linear combinations, meaning that $\mathrm{posi}(\mathcal{A}) = \mathcal{A}$.

For any two gambles $f$ and $g$ on $\mathcal{X}$, we write '$f \geq g$' if $(\forall x \in \mathcal{X}) f(x) \geq g(x)$, and '$f > g$' if $f \geq g$ and $f \neq g$. A gamble $f > 0$ is called *positive*. A gamble $g \leq 0$ is called *non-positive*. $\mathcal{G}(\mathcal{X})_{\neq 0}$ denotes the set of all non-zero gambles, $\mathcal{G}(\mathcal{X})_{>0}$ the convex cone of all positive gambles, and $\mathcal{G}(\mathcal{X})_{\leq 0}$ the convex cone of all non-positive gambles.

---

3. All the results in this section remain valid when working with more general, possibly infinite, possibility spaces, and in that case gambles are assumed to be bounded real-valued functions. We make this finiteness assumption here to avoid having to deal with the controversial issue of conglomerability (Miranda, Zaffalon, & De Cooman, 2012; Walley, 1991), because it will make the discussion of independence in later sections significantly easier, and because most practically implementable inference systems in AI are finitary in any case.

### 2.1 Coherence and Avoiding Non-positivity

**Definition 1 (Avoiding non-positivity and coherence).** *We say that a set of desirable gambles $\mathcal{D} \subseteq \mathcal{G}(\mathcal{X})$ avoids non-positivity if $f \not\leq 0$ for all gambles $f$ in $\mathrm{posi}(\mathcal{D})$, or in other words if $\mathcal{G}(\mathcal{X})_{\leq 0} \cap \mathrm{posi}(\mathcal{D}) = \emptyset$. It is called* coherent *if it satisfies the following requirements:*

D1. $0 \notin \mathcal{D}$;

D2. $\mathcal{G}(\mathcal{X})_{>0} \subseteq \mathcal{D}$;

D3. $\mathcal{D} = \mathrm{posi}(\mathcal{D})$.

*We denote by $\mathbb{D}(\mathcal{X})$ the set of all coherent sets of desirable gambles on $\mathcal{X}$.*

Requirement D3 turns $\mathcal{D}$ into a *convex cone*. Due to D2, it includes $\mathcal{G}(\mathcal{X})_{>0}$; due to D1–D3, it excludes $\mathcal{G}(\mathcal{X})_{\leq 0}$, and therefore avoids non-positivity:

D4. if $f \leq 0$ then $f \notin \mathcal{D}$, or equivalently $\mathcal{G}(\mathcal{X})_{\leq 0} \cap \mathcal{D} = \emptyset$.

The set $\mathcal{G}(\mathcal{X})_{>0}$ is coherent, and it is the smallest such subset of $\mathcal{G}(\mathcal{X})$. This set represents minimal commitments on the part of the subject, in the sense that if he knows nothing about the likelihood of the different outcomes he will only prefer to zero those gambles which are sure to never decrease his wealth and have a possibility of increasing it. Hence, it is usually taken to model complete ignorance, and it is called the *vacuous model*.

One interesting feature of coherent sets of desirable gambles is that they are linked to the field of decision making with incomplete preferences (Aumann, 1962; Dubra, Maccheroni, & Ok, 2004; Shapley & Baucells, 1998), because they are formally equivalent to the strict versions of *partial preference orderings* (Buehler, 1976; Giron & Rios, 1980). Given a coherent set of desirable gambles $\mathcal{D}$, we can define a *strict preference* relation $\succ$ between gambles by

$$f \succ g \Leftrightarrow f - g \in \mathcal{D} \text{ for any gambles } f \text{ and } g \text{ in } \mathcal{G}(\mathcal{X}).$$

Indeed, due to the linearity of the utility scale, exchanging a gamble $g$ for a gamble $f$ is a transaction with reward function $f - g$, and strictly preferring $f$ over $g$ means that this exchange should be strictly preferred to the status quo (zero). The relation $\succ$ satisfies the following conditions:

SP1. $f \not\succ f$ for all $f \in \mathcal{G}(\mathcal{X})$ [irreflexivity]

SP2. $f > g \Rightarrow f \succ g$ for all $f, g \in \mathcal{G}(\mathcal{X})$ [monotonicity]

SP3. $f \succ g$ and $g \succ h \Rightarrow f \succ h$ for all $f, g, h \in \mathcal{G}(\mathcal{X})$ [transitivity]

SP4. $f \succ g \Leftrightarrow \mu f + (1-\mu)h \succ \mu g + (1-\mu)h$ for all $\mu \in (0,1]$ and $f, g, h \in \mathcal{G}(\mathcal{X})$ [mixture independence]

Conversely, any preference relation satisfying the above axioms determines a coherent set of desirable gambles. Partial preference orderings provide a foundation for a general decision theory with imprecise probabilities and imprecise utilities (Fishburn, 1975; Seidenfeld et al., 1995; Seidenfeld, Schervish, & Kadane, 2010). See also the work by Moral and Wilson (1995), Walley (1991, 2000) and Quaeghebeur (2012b, Section 2.4) for more information.

## 2.2 Natural Extension

If we consider any non-empty family of coherent sets of desirable gambles $\mathcal{D}_i$, $i \in I$, then their intersection $\bigcap_{i \in I} \mathcal{D}_i$ is still coherent. This is the idea behind the following result, which brings to the fore a notion of coherent inference. If a subject gives us an *assessment*, a set $\mathcal{A} \subseteq \mathcal{G}(\mathcal{X})$ of gambles on $\mathcal{X}$ that he finds desirable, then it tells us exactly when this assessment can be extended to a coherent set of desirable gambles, and how to construct the smallest such set.

**Theorem 1 (De Cooman & Quaeghebeur, 2012).** *Consider $\mathcal{A} \subseteq \mathcal{G}(\mathcal{X})$, and define its* natural extension *by:*[4]

$$\mathcal{E}(\mathcal{A}) \coloneqq \bigcap \{\mathcal{D} \in \mathbb{D}(\mathcal{X}) \colon \mathcal{A} \subseteq \mathcal{D}\}.$$

*Then the following statements are equivalent:*

(i) *$\mathcal{A}$ avoids non-positivity;*

(ii) *$\mathcal{A}$ is included in some coherent set of desirable gambles;*

(iii) *$\mathcal{E}(\mathcal{A}) \neq \mathcal{G}(\mathcal{X})$;*

(iv) *the set of desirable gambles $\mathcal{E}(\mathcal{A})$ is coherent;*

(v) *$\mathcal{E}(\mathcal{A})$ is the smallest coherent set of desirable gambles that includes $\mathcal{A}$.*

*When any (and hence all) of these equivalent statements hold, then*

$$\mathcal{E}(\mathcal{A}) = \mathrm{posi}\big(\mathcal{G}(\mathcal{X})_{>0} \cup \mathcal{A}\big).$$

This shows that if we have an assessment $\mathcal{A}$ with a finite description, we can represent its natural extension on a computer by storing a finite description of its extreme rays. Although in general our assessments $\mathcal{A}$ need not have a finite description (for instance those considered in Eq. (3) further on can but need not have one), they will be of interest in a vast range of practical situations. For a description of the many cases where partial probability assessments can be given a finite description, and for efficient algorithms for verifying the coherence or computing the natural extension of a set of gambles, we refer to the work by Couso and Moral (2011) and Quaeghebeur (2012a).

## 2.3 Connection with Classical Propositional Logic

The definition of a coherent set of desirable gambles, and Theorem 1, make clear that inference with desirable gambles bears a formal resemblance to deduction in classical proposition logic: D3 is a production axiom that states that positive linear combinations of desirable gambles are again desirable. The exact correspondences are listed in the following table:

|    Classical propositional logic | Sets of desirable gambles |
|---:|:---|
| logical consistency | avoiding non-positivity |
| deductively closed | coherent |
| deductive closure | natural extension |

---

4. As usual, in this expression, we let $\bigcap \emptyset = \mathcal{G}(\mathcal{X})$.

We shall see that this inference with sets of desirable gambles has (precise-)probabilistic inference, and in particular Bayes's Rule, as a special case. But it is easy to see that it also generalises (includes as a special case) classical propositional logic: a proposition $p$ can be identified with a subset $A_p$ of the Stone space $\mathcal{X}$, and accepting a proposition $p$ corresponds to judging the gamble $\mathbb{I}_{A_p} - 1 + \epsilon$ to be desirable for all $\epsilon > 0$.[5] Here $\mathbb{I}_{A_p}$ is the so-called *indicator* (gamble) of $A_p$, assuming the value 1 on $A_p$ and 0 elsewhere. See the work by De Cooman (2005) for a more detailed discussion.

### 2.4 Helpful Lemmas

In order to prove a number of results in this paper, we need the following lemmas, one of which is convenient version of the separating hyperplane theorem. They rely heavily on the assumption of a finite space $\mathcal{X}$, and are not easily extended to a more general case.

**Lemma 2.** *Assume that $\mathcal{X}$ is finite, and consider a finite subset $\mathcal{A}$ of $\mathcal{G}(\mathcal{X})$. Then $0 \notin \mathrm{posi}(\mathcal{G}(\mathcal{X})_{>0} \cup \mathcal{A})$ if and only if there is some probability mass function $p$ such that $p(x) > 0$ for all $x \in \mathcal{X}$ and $\sum_{x \in \mathcal{X}} p(x)f(x) > 0$ for all $f \in \mathcal{A}$.*

*Proof.* It clearly suffices to prove necessity. Since $0 \notin \mathrm{posi}(\mathcal{G}(\mathcal{X})_{>0} \cup \mathcal{A})$, we infer from a version of the separating hyperplane theorem (Walley, 1991, Appendix E.1) that there is a linear functional $\Lambda$ on $\mathcal{G}(\mathcal{X})$ such that

$$(\forall x \in \mathcal{X}) \Lambda(\mathbb{I}_{\{x\}}) > 0 \text{ and } (\forall f \in \mathcal{A}) \Lambda(f) > 0.$$

Then $\Lambda(\mathcal{X}) = \sum_{x \in \mathcal{X}} \Lambda(\mathbb{I}_{\{x\}}) > 0$, and if we let $p(x) := \Lambda(\mathbb{I}_{\{x\}})/\Lambda(\mathcal{X}) > 0$ for all $x \in \mathcal{X}$, then $p$ is a probability mass function on $\mathcal{X}$ for which $\Lambda(f)/\Lambda(\mathcal{X}) = \sum_{x \in \mathcal{X}} p(x)f(x) > 0$ for all $f \in \mathcal{A}$. □

Our second lemma implies that if we consider a coherent set of desirable gambles that does not include a gamble nor its opposite, we can always find a coherent superset that includes one of the two:

**Lemma 3.** *Consider a convex cone $\mathcal{A}$ of gambles on $\mathcal{X}$ such that $\max f > 0$ for all $f \in \mathcal{A}$. Consider any non-zero gamble $g$ on $\mathcal{X}$. If $g \notin \mathcal{A}$ then $0 \notin \mathrm{posi}(\mathcal{A} \cup \{-g\})$.*

*Proof.* Consider a non-zero gamble $g \notin \mathcal{A}$, and assume *ex absurdo* that $0 \in \mathrm{posi}(\mathcal{A} \cup \{-g\})$. Then it follows from the assumptions that there are $f \in \mathcal{A}$ and $\mu > 0$ such that $0 = f + \mu(-g)$. Hence $g \in \mathcal{A}$, a contradiction. □

### 2.5 Maximal Coherent Sets of Desirable Gambles

An element $\mathcal{D}$ of $\mathbb{D}(\mathcal{X})$ is called *maximal* if it is not strictly included in any other element of $\mathbb{D}(\mathcal{X})$, or in other words, if adding any gamble $f$ to $\mathcal{D}$ makes sure we can no longer extend the set $\mathcal{D} \cup \{f\}$ to a set that is still coherent:

$$(\forall \mathcal{D}' \in \mathbb{D}(\mathcal{X}))(\mathcal{D} \subseteq \mathcal{D}' \Rightarrow \mathcal{D} = \mathcal{D}').$$

---

5. This is not equivalent to judging the gamble $\mathbb{I}_{A_p} - 1$ to be desirable, as in that case we do not obtain a coherent set of desirable gambles; the gamble $\mathbb{I}_{A_p} - 1$ is only *almost*-desirable in the sense of Walley (1991, Section 3.7.3).

$\mathbb{M}(\mathcal{X})$ denotes the set of all maximal elements of $\mathbb{D}(\mathcal{X})$.

The following proposition provides a useful characterisation of such maximal elements.

**Proposition 4 (De Cooman & Quaeghebeur, 2012).** *Consider any $\mathcal{D} \in \mathbb{D}(\mathcal{X})$. It is a maximal coherent set of desirable gambles if and only if*

$$(\forall f \in \mathcal{G}(\mathcal{X})_{\neq 0})(f \notin \mathcal{D} \Rightarrow -f \in \mathcal{D}).$$

As is the case for classical propositional logic (see, for instance, De Cooman, 2005), coherence and inference can be described completely in terms of such maximal elements. This is the essence of the following important result, which continues to hold for infinite $\mathcal{X}$, but for which a constructive proof can be given in case $\mathcal{X}$ is finite, based on the argument suggested by Couso and Moral (2011).

**Theorem 5 (Couso & Moral, 2011; De Cooman & Quaeghebeur, 2012).** *A set $\mathcal{A}$ avoids non-positivity if and only if there is some maximal $\mathcal{M} \in \mathbb{M}(\mathcal{X})$ such that $\mathcal{A} \subseteq \mathcal{M}$. Moreover*

$$\mathcal{E}(\mathcal{A}) = \bigcap m(\mathcal{A}),$$

*where we let*

$$m(\mathcal{A}) \coloneqq \{\mathcal{M} \in \mathbb{M}(\mathcal{X}) \colon \mathcal{A} \subseteq \mathcal{M}\}. \tag{1}$$

This shows that (coherent) sets of desirable gambles are instances of the so-called *strong belief structures* described and studied in detail by De Cooman (2005), into which the strong belief structures of classical propositional logic can be embedded. This guarantees amongst other things that an AGM-like (De Cooman, 2005; Gärdenfors, 1988) account of belief expansion and revision is possible for these objects.

### 2.6 Coherent Lower Previsions

We conclude this section by shedding some light on the connection between coherent sets of desirable gambles, coherent lower previsions, and probabilities.

Given a coherent set of desirable gambles $\mathcal{D}$, the functional $\underline{P}$ defined on $\mathcal{G}(\mathcal{X})$ by

$$\underline{P}(f) \coloneqq \sup\{\mu \colon f - \mu \in \mathcal{D}\} \tag{2}$$

is a *coherent lower prevision* (Walley, 1991, Thm. 3.8.1), that is, it corresponds to taking the lower envelope of the expectations associated with a set of finitely additive probabilities. The *conjugate upper prevision* $\overline{P}$ is defined by $\overline{P}(f) \coloneqq \inf\{\mu \colon \mu - f \in \mathcal{D}\} = -\underline{P}(-f)$.

Many different coherent sets of desirable gambles induce the same coherent lower prevision $\underline{P}$, and they typically differ only in their boundaries. In this sense, we can say that sets of desirable gambles are more informative than coherent lower previsions: although a gamble with positive lower prevision is always desirable and one with a negative lower prevision is not desirable, a lower prevision does not generally provide information about the desirability of a gamble whose lower prevision equal to zero. This is the reason why we need to consider the sets of desirable gambles if we want to have this additional information. To see this more clearly, consider the following adaptation of an example by Moral (2005, Example 1):

**Example 1.** *Consider $\mathcal{X}_1 = \mathcal{X}_2 = \{a,b\}$, and let $\underline{P}$ be the coherent lower prevision on $\mathcal{G}(\mathcal{X}_1 \times \mathcal{X}_2)$ given by*
$$\underline{P}(f) := \min\left\{\frac{f(b,a) + f(b,b)}{2}, \frac{f(b,a) + 3f(b,b)}{4}\right\} \text{ for all gambles } f \text{ on } \mathcal{X}_1 \times \mathcal{X}_2.$$
*This coherent lower prevision is induced by each of the following coherent sets of desirable gambles by means of Eq. (2):*
$$\mathcal{D} := \{f \colon f > 0 \text{ or } \underline{P}(f) > 0\}$$
$$\mathcal{D}' := \mathcal{D} \cup \{f \colon f(b,a) = f(b,b) = 0 \text{ and } f(a,a) + f(a,b) > 0\}.$$
*However, these two sets encode different preferences, as the gamble $g$ given by $g(a,a) = 2$, $g(a,b) = -1$, $g(b,a) = g(b,b) = 0$, with $\underline{P}(g) = 0$, is considered desirable for $\mathcal{D}'$ but not for $\mathcal{D}$. This is because coherent lower previsions are not able to distinguish between preferences and weak preferences, while sets of desirable gambles can. We shall see in Section 5 that these differences come into play when considering conditioning.* □

The smallest set of desirable gambles that induces a given coherent lower prevision—an open cone—is called the associated set of *strictly desirable* gambles, and is given by
$$\underline{\mathcal{D}} := \{f \in \mathcal{G}(\mathcal{X}) \colon f > 0 \text{ or } \underline{P}(f) > 0\}. \tag{3}$$
This is for instance the case of the set $\mathcal{D}$ in Example 1. Sets of strictly desirable gambles are in a one-to-one relationship with coherent lower previsions, and as such they suffer from the same problems when conditioning on sets of (lower) probability zero, in the sense that in the conditional models they determine in that case—by means of Eqs. (8) and (10) in Section 5—are always vacuous (Zaffalon & Miranda, 2012; Quaeghebeur, 2012b). This is one of the reasons why in this paper we are considering the more general model of coherent sets of (not necessarily strictly) desirable gambles. For additional discussion about why sets of desirable gambles are more informative than coherent lower previsions, we refer to Walley (2000) and Quaeghebeur (2012b).

When the lower and the upper prevision coincide on all gambles, then the functional $P$ defined on $\mathcal{G}(\mathcal{X})$ by $P(f) := \underline{P}(f) = \overline{P}(f)$ for all $f \in \mathcal{G}(\mathcal{X})$ is a *linear prevision*, i.e., it corresponds to the expectation operator with respect to a finitely additive probability. This happens in particular if $\mathcal{D}$ is a maximal coherent set of desirable gambles $\mathcal{M}$:
$$\underline{P}(f) = \sup\{\mu \colon f - \mu \in \mathcal{M}\} = \inf\{\mu \colon f - \mu \notin \mathcal{M}\} = \inf\{\mu \colon \mu - f \in \mathcal{M}\} = \overline{P}(f);$$
to see why the second equality holds, observe that if $f - \mu \in \mathcal{M}$ then also $f - \mu' \in \mathcal{M}$ for all $\mu' < \mu$, and as a consequence the set $\{\mu \colon f - \mu \in \mathcal{M}\}$ is an interval that is unbounded below. The third equality follows from Proposition 4. Thus, up to boundary behaviour, precise probability models correspond to a maximal coherent sets of desirable gambles; see the work by Couso and Moral (2011, Section 5), Miranda and Zaffalon (2010, Proposition 6) and Williams (1975a) for more information. Moreover, any coherent lower prevision $\underline{P}$ is the lower envelope of the *credal set* $\mathcal{M}(\underline{P})$ it induces, given by
$$\mathcal{M}(\underline{P}) := \{P \text{ linear prevision} \colon (\forall f \in \mathcal{G}(\mathcal{X}))P(f) \geq \underline{P}(f)\}.$$

We can conclude at this point that at least in its basic representational aspects, models involving coherent sets of desirable gambles generalise both classical propositional logic and precise probability in its finitary approach championed by de Finetti (1937, 1975).

## 3. Basic Notation

Now that we have highlighted the basic facts about this more general approach to uncertainty modelling, we are ready to turn to independence. In order to talk about this, we need to be able to deal with models involving more than one variable. In the present section, we introduce the notational devices we will use to make this discussion as elegant as possible.

From now on, we consider a number of variables $X_n$, $n \in N$, taking values in the respective *finite* sets $\mathcal{X}_n$. Here $N$ is some finite non-empty index set.[6]

For every subset $R$ of $N$, we denote by $X_R$ the tuple of variables (with one component for each $r \in R$) that takes values in the Cartesian product $\mathcal{X}_R \coloneqq \times_{r \in R} \mathcal{X}_r$. This Cartesian product is the set of all maps $x_R$ from $R$ to $\bigcup_{r \in R} \mathcal{X}_r$ such that $x_r \coloneqq x_R(r) \in \mathcal{X}_r$ for all $r \in R$. Elements of $\mathcal{X}_R$ are generically denoted by $x_R$ or $z_R$, with corresponding components $x_r \coloneqq x_R(r)$ or $z_r \coloneqq z_R(r)$, $r \in R$.

We will assume that the variables $X_n$ are logically independent, which means that for each subset $R$ of $N$, $X_R$ may assume all values in $\mathcal{X}_R$.

We denote by $\mathcal{G}(\mathcal{X}_R)$ the set of gambles defined on $\mathcal{X}_R$. We will frequently resort to the simplifying device of *identifying* a gamble on $\mathcal{X}_R$ with a gamble on $\mathcal{X}_N$, namely its cylindrical extension. To give an example, if $\mathcal{K} \subseteq \mathcal{G}(\mathcal{X}_N)$, this trick allows us to consider $\mathcal{K} \cap \mathcal{G}(\mathcal{X}_R)$ as the set of those gambles in $\mathcal{K}$ that depend only on the variable $X_R$. As another example, this device allows us to identify the gambles $\mathbb{I}_{\{x_R\}}$ and $\mathbb{I}_{\{x_R\} \times \mathcal{X}_{N \setminus R}}$, and therefore also the events $\{x_R\}$ and $\{x_R\} \times \mathcal{X}_{N \setminus R}$. More generally, for any event $A \subseteq \mathcal{X}_R$, we can identify the gambles $\mathbb{I}_A$ and $\mathbb{I}_{A \times \mathcal{X}_{N \setminus R}}$, and therefore also the events $A$ and $A \times \mathcal{X}_{N \setminus R}$.

We must pay particular attention to the case $R = \emptyset$. By definition, $\mathcal{X}_\emptyset$ is the set of all maps from $\emptyset$ to $\bigcup_{r \in \emptyset} \mathcal{X}_r = \emptyset$. It contains only one element $x_\emptyset$: the empty map. This means that there is no uncertainty about the value of the variable $X_\emptyset$: it can assume only one value (the empty map). Moreover $\mathbb{I}_{\mathcal{X}_\emptyset} = \mathbb{I}_{\{x_\emptyset\}} = 1$. Also, we can identify $\mathcal{G}(\mathcal{X}_\emptyset)$ with the set of real numbers $\mathbb{R}$. There is only one coherent set of desirable gambles on $\mathcal{X}_\emptyset$: the set $\mathbb{R}_{>0}$ of positive real numbers.

One final notational convention that is very handy and will be used throughout: if $n$ is an index, then we identify $n$ and $\{n\}$. So we take $\mathcal{X}_{\{n\}}$, $\mathcal{G}(\mathcal{X}_{\{n\}})$, $\mathcal{D}_{\{n\}}$ to also refer to $\mathcal{X}_n$, $\mathcal{G}(\mathcal{X}_n)$ and $\mathcal{D}_n$, respectively. This trick, amongst other things, allows us to consider two disjoint index sets $N_1$ and $N_2$, and consider each of these sets to constitute an index in themselves, leading to a new index set $\{N_1, N_2\}$. The variables $X_{N_1}$ and $X_{N_2}$ can then be combined into a joint variable $X_{\{N_1, N_2\}}$, which can of course be identified with the variable $X_{N_1 \cup N_2}$: joint variables can be considered as single variables, and combined to constitute new joint variables.

## 4. Marginalisation and Cylindrical Extension

Suppose that we have a set $\mathcal{D}_N \subseteq \mathcal{G}(\mathcal{X}_N)$ of desirable gambles modelling a subject's information about the uncertain variable $X_N$.

---

6. The assumption of finiteness of the spaces $\mathcal{X}_n$ is essential for the proofs of some of the results established later on, such as Theorem 13 and Proposition 18. It also allows us to simplify some of the expressions of the sets of gambles derived by an assumption of epistemic irrelevance or independence, from which we derive for instance Lemma 11 and Proposition 14.

Irrelevant and Independent Natural Extension for Sets of Desirable Gambles

We are interested in modelling the information about the variable $X_O$, where $O$ is some subset of $N$. This can be done using the set of desirable gambles that belong to $\mathcal{D}_N$ but only depend on the variable $X_O$:

$$\mathrm{marg}_O(\mathcal{D}_N) \coloneqq \{g \in \mathcal{G}(\mathcal{X}_O) \colon g \in \mathcal{D}_N\} = \mathcal{D}_N \cap \mathcal{G}(\mathcal{X}_O). \tag{4}$$

Observe that if $\mathcal{D}_N$ is coherent we obtain $\mathrm{marg}_\emptyset(\mathcal{D}_N) = \mathcal{G}(\mathcal{X}_\emptyset)_{>0}$, which can be identified with the set of positive real numbers $\mathbb{R}_{>0}$. Also, with $O_1 \subseteq O_2 \subseteq N$:

$$\begin{aligned}
\mathrm{marg}_{O_1}(\mathrm{marg}_{O_2}(\mathcal{D}_N)) &= \{g \in \mathcal{G}(\mathcal{X}_{O_1}) \colon g \in \mathrm{marg}_{O_2}(\mathcal{D}_N)\} \\
&= \{g \in \mathcal{G}(\mathcal{X}_{O_1}) \colon g \in \mathcal{G}(\mathcal{X}_{O_2}) \cap \mathcal{D}_N\} \\
&= \{g \in \mathcal{G}(\mathcal{X}_{O_1}) \colon g \in \mathcal{D}_N\} = \mathrm{marg}_{O_1}(\mathcal{D}_N). 
\end{aligned} \tag{5}$$

Coherence is trivially preserved under marginalisation.

**Proposition 6.** *Let $\mathcal{D}_N$ be a set of desirable gambles on $\mathcal{X}_N$, and consider any subset $O$ of $N$.*

(i) *If $\mathcal{D}_N$ avoids non-positivity, then so does $\mathrm{marg}_O(\mathcal{D}_N)$.*

(ii) *If $\mathcal{D}_N$ is coherent, then $\mathrm{marg}_O(\mathcal{D}_N)$ is a coherent set of desirable gambles on $\mathcal{X}_O$.*

We now look for a kind of inverse operation to marginalisation. Suppose we have a coherent set of desirable gambles $\mathcal{D}_O \subseteq \mathcal{G}(\mathcal{X}_O)$ modelling a subject's information about the uncertain variable $X_O$, and we want to extend this to a coherent set of desirable gambles on $\mathcal{X}_N$, representing the same information. So we are looking for a coherent set of desirable gambles $\mathcal{D}_N \subseteq \mathcal{G}(\mathcal{X}_N)$ such that $\mathrm{marg}_O(\mathcal{D}_N) = \mathcal{D}_O$ and that is *as small as possible*: the most conservative coherent set of desirable gambles on $\mathcal{X}_N$ that marginalises to $\mathcal{D}_O$. It turns out that such a set always exists and is not difficult to find.

**Proposition 7.** *Let $O$ be a subset of $N$ and let $\mathcal{D}_O \in \mathbb{D}(\mathcal{X}_O)$. Then the most conservative (smallest) coherent set of desirable gambles on $\mathcal{X}_N$ that marginalises to $\mathcal{D}_O$ is given by*

$$\mathrm{ext}_N(\mathcal{D}_O) \coloneqq \mathrm{posi}(\mathcal{G}(\mathcal{X}_N)_{>0} \cup \mathcal{D}_O). \tag{6}$$

*It is called the* cylindrical extension *of $\mathcal{D}_O$ to a set of desirable gambles on $\mathcal{X}_N$, and clearly satisfies*

$$\mathrm{marg}_O(\mathrm{ext}_N(\mathcal{D}_O)) = \mathcal{D}_O. \tag{7}$$

This extension is called *weak extension* by Moral (2005, Section 2.1).[7]

*Proof.* It is clear from the coherence requirements and Eq. (4) that any coherent set of desirable gambles that marginalises to $\mathcal{D}_O$ must include $\mathcal{G}(\mathcal{X}_N)_{>0}$ and $\mathcal{D}_O$, and therefore also $\mathrm{posi}(\mathcal{G}(\mathcal{X}_N)_{>0} \cup \mathcal{D}_O) = \mathrm{ext}_N(\mathcal{D}_O)$. It therefore suffices to prove that $\mathrm{posi}(\mathcal{G}(\mathcal{X}_N)_{>0} \cup \mathcal{D}_O)$ is coherent, and that it marginalises to $\mathcal{D}_O$.

---

7. The main difference between our result and Moral's is that we are excluding the zero gamble from any coherent set of desirable gambles, while Moral is including it.

To prove coherence, it suffices to prove that $\mathcal{D}_O$ avoids non-positivity, by Theorem 1. But this is obvious because $\mathcal{D}_O$ is a coherent set of desirable gambles on $\mathcal{X}_O$.

We are left to prove that $\mathrm{marg}_O(\mathrm{ext}_N(\mathcal{D}_O)) = \mathcal{D}_O$. Since for any $g \in \mathcal{D}_O$ it is obvious that both $g \in \mathrm{ext}_N(\mathcal{D}_O)$ and $g \in \mathcal{G}(\mathcal{X}_O)$, we see immediately that $\mathcal{D}_O \subseteq \mathrm{marg}_O(\mathrm{ext}_N(\mathcal{D}_O))$, so we concentrate on proving the converse inclusion. Consider $f \in \mathrm{marg}_O(\mathrm{ext}_N(\mathcal{D}_O))$, meaning that both $f \in \mathcal{G}(\mathcal{X}_O)$ and $f \in \mathrm{ext}_N(\mathcal{D}_O)$. The latter means that there are $g \in \mathcal{D}_O$, $h \in \mathcal{G}(\mathcal{X}_N)_{>0}$, and non-negative $\lambda$ and $\mu$ such that $\max\{\lambda, \mu\} > 0$ for which $f = \lambda g + \mu h$. Since we need to prove that $f \in \mathcal{D}_O$, we can assume without loss of generality that $\mu > 0$. But then $h = (f - \lambda g)/\mu \in \mathcal{G}(\mathcal{X}_O)$ and therefore also $h \in \mathcal{G}(\mathcal{X}_O)_{>0}$, whence indeed $f \in \mathcal{D}_O$, by coherence of $\mathcal{D}_O$. □

## 5. Conditioning

Suppose that we have a set $\mathcal{D}_N \subseteq \mathcal{G}(\mathcal{X}_N)$ of desirable gambles modelling a subject's information about the uncertain variable $X_N$.

Consider a subset $I$ of $N$, and assume we want to update the model $\mathcal{D}_N$ with the information that $X_I = x_I$. This leads to an *updated*, or *conditioned*, set of desirable gambles:

$$\mathcal{D}_N | x_I := \left\{ f \in \mathcal{G}(\mathcal{X}_N) : f > 0 \text{ or } \mathbb{I}_{\{x_I\}} f \in \mathcal{D}_N \right\}. \tag{8}$$

For technical reasons, and mainly in order to streamline the proofs as much as possible, we also allow the admittedly pathological case that $I = \emptyset$. Since $\mathbb{I}_{\{x_\emptyset\}} = 1$, this amounts to not conditioning at all.

Eq. (8) introduces the conditioning operator '|' essentially used by Walley (2000) and Moral (2005). We prefer the slightly modified version '⌋', introduced by De Cooman and Quaeghebeur (2012). Since $\mathbb{I}_{\{x_I\}} f = \mathbb{I}_{\{x_I\}} f(x_I, \cdot)$, we can characterise the updated model $\mathcal{D}_N | x_I$ through the set

$$\mathcal{D}_N \rfloor x_I := \left\{ g \in \mathcal{G}(\mathcal{X}_{N \setminus I}) : \mathbb{I}_{\{x_I\}} g \in \mathcal{D}_N \right\} \subseteq \mathcal{G}(\mathcal{X}_{N \setminus I}),$$

in the specific sense that for all $g \in \mathcal{G}(\mathcal{X}_{N \setminus I})$:

$$g \in \mathcal{D}_N \rfloor x_I \Leftrightarrow \mathbb{I}_{\{x_I\}} g \in \mathcal{D}_N \Leftrightarrow \mathbb{I}_{\{x_I\}} g \in \mathcal{D}_N | x_I, \tag{9}$$

and for all $f \in \mathcal{G}(\mathcal{X}_N)$:

$$f \in \mathcal{D}_N | x_I \Leftrightarrow (f > 0 \text{ or } f(x_I, \cdot) \in \mathcal{D}_N \rfloor x_I).$$

As the above equation shows, there is a one-to-one correspondence between $\mathcal{D}_N | x_I$ and $\mathcal{D}_N \rfloor x_I$. We prefer this second operator because we find it more intuitive that conditioning a gamble on some $x_I \in \mathcal{X}_I$ produces a gamble that depends only on the remaining $N \setminus I$ variables. This will be useful for instance when combining conditional sets of gambles, as in Proposition 24 later on.

It is immediate to prove that conditioning preserves coherence:

**Proposition 8.** *Let $\mathcal{D}_N$ be a coherent set of desirable gambles on $\mathcal{X}_N$, and consider any subset $I$ of $N$. Then $\mathcal{D}_N \rfloor x_I$ is a coherent set of desirable gambles on $\mathcal{X}_{N \setminus I}$.*

The order of marginalisation and conditioning can be reversed, under some conditions:

**Proposition 9.** *Let $\mathcal{D}_N$ be a coherent set of desirable gambles on $\mathcal{X}_N$, and consider any disjoint subsets $I$ and $O$ of $N$. Then for all $x_I \in \mathcal{X}_I$:*

$$\mathrm{marg}_O(\mathcal{D}_N\rfloor x_I) = \mathrm{marg}_{I \cup O}(\mathcal{D}_N)\rfloor x_I.$$

*Proof.* Consider any $h \in \mathcal{G}(\mathcal{X}_N)$ and observe the following chain of equivalences:

$$\begin{aligned}
h \in \mathrm{marg}_O(\mathcal{D}_N\rfloor x_I) &\Leftrightarrow h \in \mathcal{G}(\mathcal{X}_O) \text{ and } h \in \mathcal{D}_N\rfloor x_I \\
&\Leftrightarrow h \in \mathcal{G}(\mathcal{X}_O) \text{ and } \mathbb{I}_{\{x_I\}} h \in \mathcal{D}_N \\
&\Leftrightarrow h \in \mathcal{G}(\mathcal{X}_O) \text{ and } \mathbb{I}_{\{x_I\}} h \in \mathrm{marg}_{I \cup O}(\mathcal{D}_N) \\
&\Leftrightarrow h \in \mathcal{G}(\mathcal{X}_O) \text{ and } h \in \mathrm{marg}_{I \cup O}(\mathcal{D}_N)\rfloor x_I \\
&\Leftrightarrow h \in \mathrm{marg}_{I \cup O}(\mathcal{D}_N)\rfloor x_I. \qquad \square
\end{aligned}$$

To end this section, let us briefly look at the consequences of this type of updating for the coherent lower previsions associated with coherent sets of desirable gambles. This will allow us to further back our claim that standard probability theory can be recovered as a special case of the theory of coherent sets of desirable gambles, as it also allows us to derive Bayes's Rule.

Let us denote by $\underline{P}_N$ the lower prevision induced by the joint model $\mathcal{D}_N$, and by $\underline{P}(\cdot|x_I)$ the conditional lower prevision on $\mathcal{G}(\mathcal{X}_{N \setminus I})$ induced by the updated set $\mathcal{D}_N\rfloor x_I$. Then for any gamble $g$ on $\mathcal{X}_{N \setminus I}$:

$$\underline{P}(g|x_I) = \sup\{\mu \colon g - \mu \in \mathcal{D}_N\rfloor x_I\} = \sup\{\mu \colon \mathbb{I}_{\{x_I\}}[g - \mu] \in \mathcal{D}_N\}. \tag{10}$$

This allows us to clarify further that sets of desirable gambles are indeed more informative than coherent lower previsions, again using the example by Moral (2005):

**Example 2.** *Consider again the lower prevision $\underline{P}$ and the coherent sets of desirable gambles $\mathcal{D}$ and $\mathcal{D}'$ from Example 1. Consider the event that $X_1 = a$, which has (upper) probability zero in both $\mathcal{D}$ and $\mathcal{D}'$. When conditioning on this event, we obtain two different updated sets: on the one hand,*

$$\mathcal{D}\rfloor(X_1 = a) = \{g \in \mathcal{G}(\mathcal{X}_2) \colon g > 0\} = \mathcal{G}(\mathcal{X}_2)_{>0}$$

*whereas*

$$\mathcal{D}'\rfloor(X_1 = a) = \{g \in \mathcal{G}(\mathcal{X}_2) \colon g(a) + g(b) > 0\}.$$

*This means that there sets represent different information when conditioning on the event of probability zero $X_1 = a$. Indeed, if we apply Eq. (10) we see that the first one induces the vacuous lower prevision $\underline{P}(g|X_1 = a) = \min\{g(a), g(b)\}$ for any gamble $g$ on $\mathcal{X}_2$, while the second one induces the uniform linear prevision: $\underline{P}(g|X_1 = a) = \frac{g(a)+g(b)}{2}$.* $\square$

The lower previsions $\underline{P}_N$ and $\underline{P}(\cdot|x_I)$ then satisfy a condition called the *Generalised Bayes Rule* (this follows from Williams, 1975b and Miranda & Zaffalon, 2010, Thm. 8):

$$\underline{P}_N(\mathbb{I}_{\{x_I\}}[g - \underline{P}(g|x_I)]) = 0. \tag{11}$$

We refer to the work by Walley (1991, 2000) for more information about this rule. It leads to Bayes's Rule in the special case that the joint model $\mathcal{D}_N$ induces a precise prevision $P_N$. Indeed, if we let $g = \mathbb{I}_{\{x_{N \setminus I}\}}$ and generically denote probability mass by $p$, we infer from Eq. (11) and the linearity of $P_N$ that $P_N(\mathbb{I}_{\{x_I\}}\mathbb{I}_{\{x_{N \setminus I}\}}) = \underline{P}(\mathbb{I}_{\{x_{N \setminus I}\}}|x_I)P_N(\mathbb{I}_{\{x_I\}})$, or in other words that $p(x_N) = p(x_{N \setminus I}|x_I)p(x_I)$. See Section 2.6 for more details on the relationship between coherent lower (and linear) previsions and sets of desirable gambles.

**Remark 1.** *A lower prevision $\underline{P}$ is also in a one-to-one correspondence with a so-called set of* almost desirable *gambles, namely*

$$\overline{\mathcal{D}} := \{f \colon \underline{P}(f) \geq 0\}.$$

*This set corresponds to the uniform closure of any coherent set of desirable gambles $\mathcal{D}$ that induces $\underline{P}$ by means of Eq. (2). Although sets of almost-desirable gambles are interesting, and allow us work with non-strict preference relations (Walley, 1991, Section 3.7.6), we have opted for considering the more general model of coherent sets of desirable gambles for two (admittedly related) reasons. Like sets of strictly desirable gambles, sets of almost-desirable gambles do not permit to elicit boundary behaviour, which may be important when updating, as we have discussed in Example 2. Moreover, conditioning a set of almost desirable gambles may produce incoherent models when sets of probability zero are involved (Miranda & Zaffalon, 2010, Proposition 5 and Example 2): if we take for instance $\mathcal{X}_1 = \mathcal{X}_2 = \{0,1\}$ and the linear prevision $P$ with mass function $p(0, \cdot) = 0$ and $p(\cdot, 1) = \frac{1}{2}$, then its associated set of almost desirable gambles is:*

$$\overline{\mathcal{D}} = \{f \colon f(1,0) + f(1,1) \geq 0\},$$

*and if we use Eq. (8) to condition this set $\overline{\mathcal{D}}$ on $X_1 = 0$, we get $\mathcal{G}(\mathcal{X}_{1,2})$, which is an incoherent set. This means that for such sets of almost desirable gambles, and more generally for sets of gambles associated with non-strict preferences, the conditioning operation must be modified in order to avoid producing incoherences. It turns out there is no unique way of doing this; see the work by Hermans (2012) for more details.* □

## 6. Irrelevant Natural Extension

We are now ready to look at the simplest type of irrelevance judgement.

**Definition 2.** *Consider two disjoint subsets $I$ and $O$ of $N$. We say that $X_I$ is* epistemically irrelevant *to $X_O$ when learning the value of $X_I$ does not influence or change our subject's beliefs about $X_O$.*

When does a set $\mathcal{D}_N$ of desirable gambles on $\mathcal{X}_N$ capture this type of epistemic irrelevance? Observing that $X_I = x_I$ turns $\mathcal{D}_N$ into the updated set $\mathcal{D}_N \rfloor x_I$ of desirable gambles on $\mathcal{X}_{N \setminus I}$, we come to the following definition:

**Definition 3.** *A set $\mathcal{D}_N$ of desirable gambles on $\mathcal{X}_N$ is said to satisfy* epistemic irrelevance *of $X_I$ to $X_O$ when*

$$\mathrm{marg}_O(\mathcal{D}_N \rfloor x_I) = \mathrm{marg}_O(\mathcal{D}_N) \text{ for all } x_I \in \mathcal{X}_I. \tag{12}$$

As before, for technical reasons we also allow $I$ and $O$ to be empty. It is clear from the definition above that the 'variable' $X_\emptyset$, about whose constant value we are certain, is epistemically irrelevant to any variable $X_O$. Similarly, we see that any variable $X_I$ is epistemically irrelevant to the 'variable' $X_\emptyset$. This seems to be in accordance with intuition. We refer to Levi (1980) and Walley (1982, 1991) for related notions in terms of coherent lower previsions or credal sets.

The epistemic irrelevance condition can be reformulated trivially in an interesting and slightly different manner.

**Proposition 10.** *Let $\mathcal{D}_N$ be a coherent set of desirable gambles on $\mathcal{X}_N$, and let $I$ and $O$ be any disjoint subsets of $N$. Then the following statements are equivalent:*

(i) $\mathrm{marg}_O(\mathcal{D}_N \rfloor x_I) = \mathrm{marg}_O(\mathcal{D}_N)$ *for all* $x_I \in \mathcal{X}_I$;

(ii) *for all* $f \in \mathcal{G}(\mathcal{X}_O)$ *and all* $x_I \in \mathcal{X}_I$: $f \in \mathcal{D}_N \Leftrightarrow \mathbb{I}_{\{x_I\}} f \in \mathcal{D}_N$.

*Proof.* It suffices to take into account that $f \in \mathrm{marg}_O(\mathcal{D}_N)$ if and only if $f \in \mathcal{D}_N$ and $f \in \mathcal{G}(\mathcal{X}_O)$, while $f \in \mathrm{marg}_O(\mathcal{D}_N \rfloor x_I)$ if and only if $f \in \mathcal{G}(\mathcal{X}_O)$ and $\mathbb{I}_{\{x_I\}} f \in \mathcal{D}_N$. □

Irrelevance assessments are most useful in constructing sets of desirable gambles from other ones. Suppose we have a coherent set $\mathcal{D}_O$ of desirable gambles on $\mathcal{X}_O$, and an assessment that $X_I$ is epistemically irrelevant to $X_O$, where $I$ and $O$ are disjoint index sets. Then how can we combine $\mathcal{D}_O$ and this structural irrelevance assessment into a coherent set of desirable gambles on $\mathcal{X}_{I \cup O}$, or more generally, on $\mathcal{X}_N$, where $N \supseteq I \cup O$? To see how this can be done in a way that is as conservative as possible, we introduce the following sets

$$\mathcal{A}_{I \to O}^{\mathrm{irr}} := \mathrm{posi}\left(\{\mathbb{I}_{\{x_I\}} g \colon g \in \mathcal{D}_O \text{ and } x_I \in \mathcal{X}_I\}\right) \tag{13}$$

$$= \{h \in \mathcal{G}(\mathcal{X}_{I \cup O})_{\neq 0} \colon (\forall x_I \in \mathcal{X}_I) h(x_I, \cdot) \in \mathcal{D}_O \cup \{0\}\}. \tag{14}$$

Clearly, and this will be quite important in streamlining proofs, $\mathcal{A}_{\emptyset \to O}^{\mathrm{irr}} = \mathcal{D}_O$ and $\mathcal{A}_{I \to \emptyset}^{\mathrm{irr}} = \mathcal{G}(\mathcal{X}_I)_{>0}$. The intuition behind Eq. (13) is to consider the cylindrical extensions of the gambles in $\mathcal{D}_O$ to the space $\mathcal{X}_{I \cup O}$, and to take the natural extension of the resulting set. The alternative expression (14) shows that this is equivalent to selecting a gamble in $\mathcal{D}_O$ for a finite number of $x_I$ in $\mathcal{X}_I$, and to derive from them a gamble on $\mathcal{X}_{I \cup O}$.

Let us give two important properties of these sets:

**Lemma 11.** *Consider disjoint subsets $I$ and $O$ of $N$, and a coherent set $\mathcal{D}_O$ of desirable gambles on $\mathcal{X}_O$. Then $\mathcal{A}_{I \to O}^{\mathrm{irr}}$ is a coherent set of desirable gambles on $\mathcal{X}_{I \cup O}$.*

*Proof.* D1. Assume *ex absurdo* that there are $n > 0$, real $\lambda_k > 0$ and $f_k \in \mathcal{A}_{I \to O}^{\mathrm{irr}}$ such that $\sum_{k=1}^n \lambda_k f_k = 0$. It follows from the assumptions that there are $\ell \in \{1, \ldots, n\}$ and $x_I \in \mathcal{X}_I$ such that $f_\ell(x_I, \cdot) \neq 0$. This implies that in the sum $\sum_{k=1}^n \lambda_k f_k(x_I, \cdot) = 0$ not all the gambles $\lambda_k f_k(x_I, \cdot)$ are zero. Since the non-zero ones belong to $\mathcal{D}_O$, this contradicts the coherence of $\mathcal{D}_O$.

D2. Consider any $h \in \mathcal{G}(\mathcal{X}_{I \cup O})_{>0}$. Then clearly $h(x_I, \cdot) \geq 0$ and therefore $h(x_I, \cdot) \in \mathcal{D}_O \cup \{0\}$ for all $x_I \in \mathcal{X}_I$. Since $h \neq 0$, it follows that indeed $h \in \mathcal{A}_{I \to O}^{\mathrm{irr}}$.

D3. Trivial, using that $\mathrm{posi}(\mathrm{posi}(\mathcal{D})) = \mathrm{posi}(\mathcal{D})$ for any set of desirable gambles $\mathcal{D}$. □

**Lemma 12.** *Consider disjoint subsets $I$ and $O$ of $N$, and a coherent set $\mathcal{D}_O$ of desirable gambles on $\mathcal{X}_O$. Then $\mathrm{marg}_O(\mathcal{A}^{\mathrm{irr}}_{I \to O}) = \mathcal{D}_O$.*

*Proof.* It is obvious from Eq. (14) that indeed:

$$\mathrm{marg}_O(\mathcal{A}^{\mathrm{irr}}_{I \to O}) = \mathcal{A}^{\mathrm{irr}}_{I \to O} \cap \mathcal{G}(\mathcal{X}_O) = \{h \in \mathcal{G}(\mathcal{X}_O)_{\neq 0} \colon (\forall x_I \in \mathcal{X}_I) h \in \mathcal{D}_O \cup \{0\}\}$$
$$= \{h \in \mathcal{G}(\mathcal{X}_O)_{\neq 0} \colon h \in \mathcal{D}_O \cup \{0\}\} = \mathcal{D}_O. \qquad \square$$

**Theorem 13.** *Consider disjoint subsets $I$ and $O$ of $N$, and a coherent set $\mathcal{D}_O$ of desirable gambles on $\mathcal{X}_O$. Then the smallest coherent set of desirable gambles on $\mathcal{X}_N$ that marginalises to $\mathcal{D}_O$ and satisfies the epistemic irrelevance condition (12) of $X_I$ to $X_O$ is given by $\mathrm{ext}_N(\mathcal{A}^{\mathrm{irr}}_{I \to O}) = \mathrm{posi}(\mathcal{G}(\mathcal{X}_N)_{>0} \cup \mathcal{A}^{\mathrm{irr}}_{I \to O})$.*

*Proof.* Consider any coherent set $\mathcal{D}_N$ of desirable gambles on $\mathcal{X}_N$ that marginalises to $\mathcal{D}_O$ and satisfies the irrelevance condition (12). This implies that $\mathrm{marg}_O(\mathcal{D}_N \rfloor x_I) = \mathcal{D}_O$ for any $x_I \in \mathcal{X}_I$, so $g \in \mathcal{D}_N \rfloor x_I$, and therefore $\mathbb{I}_{\{x_I\}} g \in \mathcal{D}_N$ for any $g \in \mathcal{D}_O$, by Eq. (9). So we infer by coherence that $\mathcal{A}^{\mathrm{irr}}_{I \to O} \subseteq \mathcal{D}_N$, and therefore also that $\mathrm{posi}(\mathcal{G}(\mathcal{X}_N)_{>0} \cup \mathcal{A}^{\mathrm{irr}}_{I \to O}) \subseteq \mathcal{D}_N$. As a consequence, it suffices to prove that (i) $\mathrm{ext}_N(\mathcal{A}^{\mathrm{irr}}_{I \to O})$ is coherent, (ii) marginalises to $\mathcal{D}_O$, and (iii) satisfies the epistemic irrelevance condition (12). This is what we now set out to do.

(i). By Lemma 11, $\mathcal{A}^{\mathrm{irr}}_{I \to O}$ is a coherent set of desirable gambles on $\mathcal{X}_{I \cup O}$, so Proposition 7 implies that $\mathrm{posi}(\mathcal{G}(\mathcal{X}_N)_{>0} \cup \mathcal{A}^{\mathrm{irr}}_{I \to O}) = \mathrm{ext}_N(\mathcal{A}^{\mathrm{irr}}_{I \to O})$ is a coherent set of desirable gambles on $\mathcal{X}_N$.

(ii). Marginalisation leads to:

$$\mathrm{marg}_O(\mathrm{ext}_N(\mathcal{A}^{\mathrm{irr}}_{I \to O})) = \mathrm{marg}_O(\mathrm{marg}_{I \cup O}(\mathrm{ext}_N(\mathcal{A}^{\mathrm{irr}}_{I \to O}))) = \mathrm{marg}_O(\mathcal{A}^{\mathrm{irr}}_{I \to O}) = \mathcal{D}_O,$$

where the first equality follows from Eq. (5), the second from Eq. (7), and the third from Lemma 12.

(iii). It follows from Proposition 9 and Eq. (7) that

$$\mathrm{marg}_O(\mathrm{ext}_N(\mathcal{A}^{\mathrm{irr}}_{I \to O}) \rfloor x_I) = \mathrm{marg}_{I \cup O}(\mathrm{ext}_N(\mathcal{A}^{\mathrm{irr}}_{I \to O})) \rfloor x_I = \mathcal{A}^{\mathrm{irr}}_{I \to O} \rfloor x_I,$$

and we have just shown in (ii) that $\mathrm{marg}_O(\mathrm{ext}_N(\mathcal{A}^{\mathrm{irr}}_{I \to O})) = \mathcal{D}_O$, so proving the equality $\mathrm{marg}_O(\mathrm{ext}_N(\mathcal{A}^{\mathrm{irr}}_{I \to O}) \rfloor x_I) = \mathrm{marg}_O(\mathrm{ext}_N(\mathcal{A}^{\mathrm{irr}}_{I \to O}))$ amounts to proving that $\mathcal{A}^{\mathrm{irr}}_{I \to O} \rfloor x_I = \mathcal{D}_O$. It is obvious from the definition of $\mathcal{A}^{\mathrm{irr}}_{I \to O}$ that $\mathcal{D}_O \subseteq \mathcal{A}^{\mathrm{irr}}_{I \to O} \rfloor x_I$, so we concentrate on the converse inclusion. Consider any $h \in \mathcal{A}^{\mathrm{irr}}_{I \to O} \rfloor x_I$; then $\mathbb{I}_{\{x_I\}} h \in \mathcal{A}^{\mathrm{irr}}_{I \to O}$, so we infer from Eq. (14) that in particular $h \in \mathcal{D}_O \cup \{0\}$. But since $\mathcal{A}^{\mathrm{irr}}_{I \to O}$ is coherent by Lemma 11, we see that $h \neq 0$ and therefore indeed $h \in \mathcal{D}_O$. $\qquad \square$

Theorem 13 is mentioned briefly, with only a hint at the proof, by Moral (2005, Section 2.4). We believe the result is not so trivial and have therefore decided to include our version of the proof here. Our notion of epistemic irrelevance is called *weak* epistemic irrelevance by Moral. For his version of epistemic irrelevance he requires in addition that $\mathcal{D}_N$ should be equal to the irrelevant natural extension of $\mathcal{D}_O$, and therefore be the *smallest* model that satisfies the (weak) epistemic irrelevance condition (12). While we feel comfortable with his reasons for doing so, we have decided not to follow his lead in this. Our main reason for not doing so is tied up with the philosophy behind partial assessments (or probability specifications). Each such assessment, be it *local* (e.g. stating that all gambles in some set $\mathcal{A}$ are desirable) or *structural* (e.g. imposing symmetry or irrelevance), serves to further restrict the possible models, and at each stage the most conservative (smallest possible) model is considered to be the one to be used, and possibly further refined by additional assessments. Only calling a model irrelevant when it is the smallest weakly irrelevant model would, we believe, conflict with approach: larger models obtained later on by adding, say, further symmetry assessments, would no longer deserve to be called irrelevant (but would still satisfy all the relevant conditions).

We infer from Theorem 13 and Eq. (13) that extreme rays of the irrelevant natural extension have the form $\mathbb{I}_{\{x_I\}}g$, where $g$ is some extreme ray of $\mathcal{D}_O$, so representing or finding this extension on a computer has a computational complexity that is linear in the number of extreme rays of $\mathcal{D}_O$ and linear in the number of elements of the product set $\mathcal{X}_I$—and therefore essentially exponential in the number $|I|$ of irrelevant variables $X_i$, $i \in I$. More generally, this will also be the case in the fairly general situation where $\mathcal{D}_O$ is generated by a finite number of so-called 'generalised' extreme rays, as described in detail in by Couso and Moral (2011, Section 4) and Quaeghebeur (2012a, Section 3).

## 7. Independent Natural Extension

We now turn to independence assessments, which constitute a symmetrisation of irrelevance assessments.

**Definition 4.** *We say that the variables $X_n$, $n \in N$ are* epistemically independent *when learning the values of any number of them does not influence or change our beliefs about the remaining ones: for any two disjoint subsets $I$ and $O$ of $N$, $X_I$ is epistemically irrelevant to $X_O$.*

When does a set $\mathcal{D}_N$ of desirable gambles on $\mathcal{X}_N$ capture this type of epistemic independence?

**Definition 5.** *A coherent set $\mathcal{D}_N$ of desirable gambles on $\mathcal{X}_N$ is called* independent *if*

$$\mathrm{marg}_O(\mathcal{D}_N \rfloor x_I) = \mathrm{marg}_O(\mathcal{D}_N) \text{ for all disjoint subsets } I \text{ and } O \text{ of } N, \text{ and all } x_I \in \mathcal{X}_I.$$

In this definition, we allow $I$ and $O$ to be empty too, but doing so does not lead to any substantive requirement, because the condition $\mathrm{marg}_O(\mathcal{D}_N \rfloor x_I) = \mathrm{marg}_O(\mathcal{D}_N)$ is trivially satisfied when $I$ or $O$ are empty.

Independent sets have an interesting factorisation property, which means that a product of two desirable gambles that depend on different variables should again be desirable, provided one of the gambles is positive; we refer to the work by De Cooman et al. (2011) for another paper where factorisation is considered in this somewhat unusual form. Factorisation follows from the characterisation of epistemic irrelevance we have given in Proposition 10 and the properties of coherence.

**Proposition 14 (Factorisation of independent sets).** *Let $\mathcal{D}_N$ be an independent coherent set of desirable gambles on $\mathcal{X}_N$. Then for all disjoint subsets $I$ and $O$ of $N$ and for all $f \in \mathcal{G}(\mathcal{X}_O)$:*

$$f \in \mathcal{D}_N \Leftrightarrow (\forall g \in \mathcal{G}(\mathcal{X}_I)_{>0})(fg \in \mathcal{D}_N). \tag{15}$$

*Proof.* Fix arbitrary disjoint subsets $I$ and $O$ of $N$ and any $f \in \mathcal{G}(\mathcal{X}_O)$; we show that Eq. (15) holds. The '$\Leftarrow$' part is trivial. For the '$\Rightarrow$' part, assume that $f \in \mathcal{D}_N$ and consider any $g \in \mathcal{G}(\mathcal{X}_I)_{>0}$. We have to show that $fg \in \mathcal{D}_N$. Since $g = \sum_{x_I \in \mathcal{X}_I} \mathbb{I}_{\{x_I\}} g(x_I)$, we see that $fg = \sum_{x_I \in \mathcal{X}_I} g(x_I) \mathbb{I}_{\{x_I\}} f$. Now since $f \in \text{marg}_O(\mathcal{D}_N)$, we infer from the independence of $\mathcal{D}_N$ and Proposition 10 that $f \in \mathcal{D}_N \rfloor x_I$ and therefore $\mathbb{I}_{\{x_I\}} f \in \mathcal{D}_N$ for all $x_I \in \mathcal{X}_I$. We conclude that $fg$ is a positive linear combination of elements $\mathbb{I}_{\{x_I\}} f$ of $\mathcal{D}_N$, and therefore belongs to $\mathcal{D}_N$ by coherence. $\square$

Independence assessments are useful in constructing joint sets of desirable gambles from marginal ones. Suppose we have coherent sets $\mathcal{D}_n$ of desirable gambles on $\mathcal{X}_n$, for each $n \in N$ and an assessment that the variables $X_n$, $n \in N$ are epistemically independent. Then how can we combine the $\mathcal{D}_n$ and this structural independence assessment into a coherent set of desirable gambles on $\mathcal{X}_N$ in a way that is as conservative as possible? If we call *independent product* of the $\mathcal{D}_n$ any independent $\mathcal{D}_N \in \mathbb{D}(\mathcal{X}_N)$ that marginalises to the $\mathcal{D}_n$ for all $n \in N$, this means we are looking for the smallest such independent product.

Further on, we are going to prove that such a smallest independent product always exists. Before we can do this elegantly, however, we need to do some preparatory work involving particular sets of desirable gambles that can be constructed from the $\mathcal{D}_n$. Consider, as a special case of Eq. (14), for any subset $I$ of $N$ and any $o \in N \setminus I$:

$$\mathcal{A}^{\text{irr}}_{I \to \{o\}} := \text{posi}\left(\{\mathbb{I}_{\{x_I\}} g \colon g \in \mathcal{D}_o \text{ and } x_I \in \mathcal{X}_I\}\right) \tag{16}$$

$$= \left\{h \in \mathcal{G}(\mathcal{X}_{I \cup \{o\}})_{\neq 0} \colon (\forall x_I \in \mathcal{X}_I) h(x_I, \cdot) \in \mathcal{D}_o \cup \{0\}\right\}, \tag{17}$$

and use these sets to construct the following set of gambles on $\mathcal{X}_N$:

$$\otimes_{n \in N} \mathcal{D}_n := \text{posi}\left(\mathcal{G}(\mathcal{X}_N)_{>0} \cup \bigcup_{n \in N} \mathcal{A}^{\text{irr}}_{N \setminus \{n\} \to \{n\}}\right) = \text{posi}\left(\bigcup_{n \in N} \mathcal{A}^{\text{irr}}_{N \setminus \{n\} \to \{n\}}\right), \tag{18}$$

where the second equality holds because the set $\mathcal{G}(\mathcal{X}_N)_{>0}$ is included in $\mathcal{A}^{\text{irr}}_{N \setminus \{n\} \to \{n\}}$ for every $n \in N$. The set $\otimes_{n \in N} \mathcal{D}_n$ gathers the subsets of $\mathcal{G}(\mathcal{X}_N)$ we can derive from the different $\mathcal{D}_n$ by means of an assumption of epistemic irrelevance, and considers the natural extension of their union, which is the minimal coherent superset (we shall show that it is indeed coherent in Proposition 15 below). Observe that, quite trivially, $\mathcal{A}^{\text{irr}}_{\{n\} \setminus \{n\} \to \{n\}} = \mathcal{D}_n$ and therefore $\otimes_{m \in \{n\}} \mathcal{D}_m = \mathcal{D}_n$. We now prove a number of important properties for $\otimes_{n \in N} \mathcal{D}_n$.

**Proposition 15 (Coherence).** *Let $\mathcal{D}_n$ be coherent sets of desirable gambles on $\mathcal{X}_n$, $n \in N$. Then $\otimes_{n \in N} \mathcal{D}_n$ is a coherent set of desirable gambles on $\mathcal{X}_N$.*

*Proof.* Let, for ease of notation, $\mathcal{A}_N := \bigcup_{n \in N} \mathcal{A}^{\text{irr}}_{N \setminus \{n\} \to \{n\}}$. It follows from Theorem 1 that we have to prove that $\mathcal{A}_N$ avoids non-positivity. So consider any $f \in \text{posi}(\mathcal{A}_N)$, and assume *ex absurdo* that $f \leq 0$. Then there are $\lambda_n \geq 0$ and $f_n \in \mathcal{A}^{\text{irr}}_{N \setminus \{n\} \to \{n\}}$ such that $f = \sum_{n \in N} \lambda_n f_n$ and $\max_{n \in N} \lambda_n > 0$ [recall that the $\mathcal{A}^{\text{irr}}_{N \setminus \{n\} \to \{n\}}$ are convex cones, by Lemma 11]. Fix arbitrary $m \in N$. Let

$$\mathcal{A}^N_m := \{f_m(x_{N \setminus \{m\}}, \cdot) \colon x_{N \setminus \{m\}} \in \mathcal{X}_{N \setminus \{m\}}, f_m(x_{N \setminus \{m\}}, \cdot) \neq 0\},$$

then it follows from Eq. (17) that $\mathcal{A}^N_m$ is a finite non-empty subset of $\mathcal{D}_m$, so the coherence of $\mathcal{D}_m$, Theorem 1 and Lemma 2 imply that there is some mass function $p_m$ on $\mathcal{X}_m$ with expectation operator $E_m$ such that $(\forall x_m \in \mathcal{X}_m) p_m(x_m) > 0$ and

$$(\forall x_{N \setminus \{m\}} \in \mathcal{X}_{N \setminus \{m\}})(f_m(x_{N \setminus \{m\}}, \cdot) \neq 0 \Rightarrow E_m(f_m(x_{N \setminus \{m\}}, \cdot)) > 0).$$

So if we define the gamble $g_{N \setminus \{m\}}$ on $\mathcal{X}_{N \setminus \{m\}}$ by letting

$$g_{N \setminus \{m\}}(x_{N \setminus \{m\}}) := E_m(f_m(x_{N \setminus \{m\}}, \cdot))$$

for all $x_{N \setminus \{m\}} \in \mathcal{X}_{N \setminus \{m\}}$, then $g_{N \setminus \{m\}} > 0$.

Since we can do this for all $m \in N$, we can define the mass function $p_N$ on $\mathcal{X}_N$ by letting $p_N(x_N) := \prod_{m \in N} p_m(x_m) > 0$ for all $x_N \in \mathcal{X}_N$. The corresponding expectation operator $E_N$ is of course the product operator of the marginals $E_m$. But then it follows from the reasoning and assumptions above that $E_N(f) = \sum_{m \in N} \lambda_m E_N(f_m) = \sum_{m \in N} \lambda_m E_N(g_m) > 0$, whereas $f \leq 0$ leads us to conclude that $E_N(f) \leq 0$, a contradiction. □

**Lemma 16.** *Consider any disjoint subsets $I$, $R$ and any $o \in N \setminus (I \cup R)$. Then $f(x_R, \cdot) \in \mathcal{A}^{\text{irr}}_{I \to \{o\}} \cup \{0\}$ for all $f \in \mathcal{A}^{\text{irr}}_{I \cup R \to \{o\}}$ and all $x_R \in \mathcal{X}_R$.*

*Proof.* Fix $f \in \mathcal{A}^{\text{irr}}_{I \cup R \to \{o\}}$ and $x_R \in \mathcal{X}_R$ and consider the gamble $g := f(x_R, \cdot)$ on $\mathcal{X}_{I \cup \{o\}}$. It follows from the assumptions that for all $x_I \in \mathcal{X}_I$:

$$g(x_I, \cdot) = f(x_R, x_I, \cdot) \in \mathcal{D}_o \cup \{0\},$$

whence indeed $g \in \mathcal{A}^{\text{irr}}_{I \to \{o\}} \cup \{0\}$. □

**Proposition 17 (Marginalisation).** *Consider coherent marginal sets of desirable gambles $\mathcal{D}_n$ for all $n \in N$. Let $R$ be any subset of $N$, then $\text{marg}_R(\otimes_{n \in N} \mathcal{D}_n) = \otimes_{r \in R} \mathcal{D}_r$.*

*Proof.* Since we are interpreting gambles on $\mathcal{X}_R$ as special gambles on $\mathcal{X}_N$, it is clear from Eq. (17) that for any $r \in R$, $\mathcal{A}^{\text{irr}}_{R \setminus \{r\} \to \{r\}} \subseteq \mathcal{A}^{\text{irr}}_{N \setminus \{r\} \to \{r\}}$. Eqs. (6) and (18) now tell us that $\text{ext}_N(\otimes_{r \in R} \mathcal{D}_r) \subseteq \otimes_{n \in N} \mathcal{D}_n$. If we invoke Eq. (7), this leads to

$$\otimes_{r \in R} \mathcal{D}_r = \text{marg}_R(\text{ext}_N(\otimes_{r \in R} \mathcal{D}_r)) \subseteq \text{marg}_R(\otimes_{n \in N} \mathcal{D}_n),$$

so we can concentrate on the converse inequality.

Consider therefore any $f \in \text{marg}_R(\otimes_{n \in N} \mathcal{D}_n) = (\otimes_{n \in N} \mathcal{D}_n) \cap \mathcal{G}(\mathcal{X}_R)$, and assume *ex absurdo* that $f \notin \otimes_{r \in R} \mathcal{D}_r$.

It follows from the coherence of $\otimes_{n \in N} \mathcal{D}_n$ that $f \neq 0$ [see Proposition 15]. Since $f \in \otimes_{n \in N} \mathcal{D}_n$, there are $S \subseteq N$, $f_s \in \mathcal{A}^{\text{irr}}_{N \setminus \{s\} \to \{s\}}$, $s \in S$ and $g \in \mathcal{G}(\mathcal{X}_N)$ with $g \geq 0$ such that $f = g + \sum_{s \in S} f_s$. Clearly $S \setminus R \neq \emptyset$, because $S \setminus R = \emptyset$ would imply that, with $x_{N \setminus R}$ any element of $\mathcal{X}_{N\setminus R}$, $f = f(x_{N\setminus R}, \cdot) = g(x_{N\setminus R}, \cdot) + \sum_{s\in S\cap R} f_s(x_{N\setminus R}, \cdot) \in \otimes_{r\in R}\mathcal{D}_r$, since we infer from Lemma 16 that $f_s(x_{N\setminus R}, \cdot) \in \mathcal{A}^{\mathrm{irr}}_{R\setminus\{s\}\to\{s\}} \cup \{0\}$ for all $s \in S \cap R$.

It follows from the coherence of $\otimes_{r\in R}\mathcal{D}_r$ [Proposition 15], $f \notin \otimes_{r\in R}\mathcal{D}_r$ and Lemma 3 that $0 \notin \mathrm{posi}(\{-f\} \cup \otimes_{r\in R}\mathcal{D}_r)$. Let, for ease of notation,

$$\mathcal{A}^N_{S\cap R} := \{f_s(z_{N\setminus R}, \cdot) \colon s \in S\cap R, z_{N\setminus R} \in \mathcal{X}_{N\setminus R}, f_s(z_{N\setminus R}, \cdot) \neq 0\}.$$

Then $\mathcal{A}^N_{S\cap R}$ is clearly a finite subset of $\otimes_{r\in R}\mathcal{D}_r$ [to see this, use a similar argument as above, involving Lemma 16], so we infer from Lemma 2 that there is some mass function $p_R$ on $\mathcal{X}_R$ with associated expectation operator $E_R$ such that

$$\begin{cases} (\forall x_R \in \mathcal{X}_R) p_R(x_R) > 0 \\ (\forall s \in S \cap R)(\forall z_{N\setminus R} \in \mathcal{X}_{N\setminus R}) E_R(f_s(z_{N\setminus R}, \cdot)) \geq 0 \\ E_R(f) < 0. \end{cases}$$

We then infer from $f = f(z_{N\setminus R}, \cdot) = g(z_{N\setminus R}, \cdot) + \sum_{s\in S\cap R} f_s(z_{N\setminus R}, \cdot) + \sum_{s\in S\setminus R} f_s(z_{N\setminus R}, \cdot)$ that for all $z_{N\setminus R}$ in $\mathcal{X}_{N\setminus R}$:

$$0 > E_R(f) - E_R(g(z_{N\setminus R}, \cdot)) - \sum_{s\in S\cap R} E_R(f_s(z_{N\setminus R}, \cdot))$$
$$= \sum_{s\in S\setminus R} E_R(f_s(z_{N\setminus R}, \cdot)) = \sum_{s\in S\setminus R} \sum_{x_R\in\mathcal{X}_R} p_R(x_R) f_s(z_{N\setminus R}, x_R).$$

The gambles $f_s(\cdot, x_R)$ on $\mathcal{X}_{N\setminus R}$ [where $x_R \in \mathcal{X}_R$ and $s \in S \setminus R$] can clearly not all be zero. The non-zero ones all belong to $\otimes_{s\in N\setminus R}\mathcal{D}_s$ for all $s \in N \setminus R$ and all $x_R \in \mathcal{X}_R$, by Lemma 16, so the coherence of the set of desirable gambles $\otimes_{s\in N\setminus R}\mathcal{D}_s$ [Proposition 15] guarantees that their positive linear combination $h := \sum_{s\in S\setminus R} \sum_{x_R\in\mathcal{X}_R} p_R(x_R) f_s(\cdot, x_R)$ also belongs to $\otimes_{s\in N\setminus R}\mathcal{D}_s$. This contradicts $h \leq 0$. Hence indeed $f \in \otimes_{r\in R}\mathcal{D}_r$. □

**Proposition 18 (Conditioning).** *Consider coherent marginal sets of desirable gambles $\mathcal{D}_n$ for all $n \in N$, and define $\otimes_{n\in N}\mathcal{D}_n$ by means of Eq. (18). Then $\otimes_{n\in N}\mathcal{D}_n$ is independent: for all disjoint subsets $I$ and $O$ of $N$, and all $x_I \in \mathcal{X}_I$,*

$$\mathrm{marg}_O(\otimes_{n\in N}\mathcal{D}_n \rfloor x_I) = \mathrm{marg}_O(\otimes_{n\in N}\mathcal{D}_n) = \otimes_{o\in O}\mathcal{D}_o.$$

This could probably be proved indirectly using the 'semi-graphoid' properties of conditional epistemic irrelevance, proved by Moral (2005); it appears we need reverse weak union, reverse decomposition, and contraction. Here we give a direct proof. Proposition 17 can also be seen as a special case of the present result for $I = \emptyset$.

*Proof.* Fix arbitrary disjoint subsets $I$ and $O$ of $N$, and arbitrary $x_I \in \mathcal{X}_I$. The second equality follows from Proposition 17, so we concentrate on proving that $\mathrm{marg}_O(\otimes_{n\in N}\mathcal{D}_n \rfloor x_I)$ coincides with $\otimes_{o\in O}\mathcal{D}_o$. The proof is similar to that of Proposition 17.

We first show that $\otimes_{o\in O}\mathcal{D}_o \subseteq \otimes_{n\in N}\mathcal{D}_n \rfloor x_I$. Consider any gamble $f \in \otimes_{o\in O}\mathcal{D}_o$, then we have to show that $\mathbb{I}_{\{x_I\}}f \in \otimes_{n\in N}\mathcal{D}_n$. By assumption, there are non-negative reals $\lambda_o$ and $\mu$, gambles $f_o \in \mathcal{A}^{\mathrm{irr}}_{O\setminus\{o\}\to\{o\}}$ for all $o \in O$ and $g \in \mathcal{G}(\mathcal{X}_O)_{>0}$ such that $f = \mu g + \sum_{o\in O} \lambda_o f_o$ and $\max\{\mu, \max_{o \in O} \lambda_o\} > 0$. Fix $o \in O$ and let $f'_o := \mathbb{I}_{\{x_I\}} f_o \in \mathcal{G}(\mathcal{X}_N)$. Then it follows from the definition of $\mathcal{A}^{\text{irr}}_{O \setminus \{o\} \to \{o\}}$ that $f'_o(z_{N \setminus \{o\}}, \cdot) = \mathbb{I}_{\{x_I\}}(z_I) f_o(z_{O \setminus \{o\}}, \cdot) \in \mathcal{D}_o \cup \{0\}$ for all $z_{N \setminus \{o\}} \in \mathcal{X}_{N \setminus \{o\}}$. Since $f'_o \neq 0$, the definition of $\mathcal{A}^{\text{irr}}_{N \setminus \{o\} \to \{o\}}$ tells us that $f'_o \in \mathcal{A}^{\text{irr}}_{N \setminus \{o\} \to \{o\}}$. Similarly, if we let $g' := \mathbb{I}_{\{x_I\}} g \in \mathcal{G}(\mathcal{X}_N)$, then $g' > 0$. So it follows from Eq. (18) that indeed $\mathbb{I}_{\{x_I\}} f = \mu g' + \sum_{o \in O} \lambda_o f'_o \in \otimes_{n \in N} \mathcal{D}_n$.

We now turn to the converse inclusion, $\otimes_{n \in N} \mathcal{D}_n \rfloor x_I \subseteq \otimes_{o \in O} \mathcal{D}_o$. Consider any gamble $f \in \mathcal{G}(\mathcal{X}_O)$ such that $\mathbb{I}_{\{x_I\}} f$ belongs to $\otimes_{n \in N} \mathcal{D}_n$ and assume *ex absurdo* that $f \notin \otimes_{o \in O} \mathcal{D}_o$. Let, for the sake of notational simplicity, $C := N \setminus (I \cup O)$.

It follows from the coherence of $\otimes_{n \in N} \mathcal{D}_n$ that $f \neq 0$ [see Proposition 15]. Since $\mathbb{I}_{\{x_I\}} f \in \otimes_{n \in N} \mathcal{D}_n$, there are $S \subseteq N$, $f_s \in \mathcal{A}^{\text{irr}}_{N \setminus \{s\} \to \{s\}}$, $s \in S$ and $g \in \mathcal{G}(\mathcal{X}_N)$ with $g \geq 0$ such that $\mathbb{I}_{\{x_I\}} f = g + \sum_{s \in S} f_s$. Clearly $S \setminus O \neq \emptyset$, because $S \setminus O = \emptyset$ would imply that, with $x_C$ any element of $\mathcal{X}_C$, $f = g(x_I, x_C, \cdot) + \sum_{s \in S \cap O} f_s(x_I, x_C, \cdot) \in \otimes_{o \in O} \mathcal{D}_o$, since Lemma 16 shows that $f_s(x_I, x_C, \cdot) \in \mathcal{A}^{\text{irr}}_{O \setminus \{s\} \to \{s\}}$ for all $s \in S \cap O$.

It follows from the coherence of $\otimes_{o \in O} \mathcal{D}_o$ [Proposition 15], $f \notin \otimes_{o \in O} \mathcal{D}_o$ and Lemma 3 that $0 \notin \text{posi}(\{-f\} \cup \otimes_{o \in O} \mathcal{D}_o)$. Let, for ease of notation,

$$\mathcal{A}^N_{S \cap O} := \{f_s(x_I, z_C, \cdot) : s \in S \cap O, z_C \in \mathcal{X}_C, f_s(x_I, z_C, \cdot) \neq 0\}.$$

Then $\mathcal{A}^N_{S \cap O}$ is clearly a finite subset of $\otimes_{o \in O} \mathcal{D}_o$ [to see this, use a similar argument as above, involving Lemma 16], so we infer from Lemma 2 that there is some mass function $p_O$ on $\mathcal{X}_O$ with associated expectation operator $E_O$ such that

$$\begin{cases} (\forall x_O \in \mathcal{X}_O) p_O(x_O) > 0 \\ (\forall s \in S \cap O)(\forall z_C \in \mathcal{X}_C) E_O(f_s(x_I, z_C, \cdot)) \geq 0 \\ E_O(f) < 0. \end{cases}$$

Since $f = g(x_I, z_C, \cdot) + \sum_{s \in S \cap O} f_s(x_I, z_C, \cdot) + \sum_{s \in S \setminus O} f_s(x_I, z_C, \cdot)$ for any choice of $z_C \in \mathcal{X}_C$, we see that:

$$0 > E_O(f) - E_O(g(x_I, z_C, \cdot)) - \sum_{s \in S \cap O} E_O(f_s(x_I, z_C, \cdot))$$
$$= \sum_{s \in S \setminus O} E_O(f_s(x_I, z_C, \cdot)) = \sum_{s \in S \setminus O} \sum_{x_O \in \mathcal{X}_O} p_O(x_O) f_s(x_I, z_C, x_O)).$$

Similarly, for any $z_C \in \mathcal{X}_C$ and any $z_I \in \mathcal{X}_I \setminus \{x_I\}$ we then infer from $0 = g(z_I, z_C, \cdot) + \sum_{s \in S \cap O} f_s(z_I, z_C, \cdot) + \sum_{s \in S \setminus O} f_s(z_I, z_C, \cdot)$ that:

$$0 \geq -E_O(g(z_I, z_C, \cdot)) - \sum_{s \in S \cap O} E_O(f_s(z_I, z_C, \cdot))$$
$$= \sum_{s \in S \setminus O} E_O(f_s(z_I, z_C, \cdot)) = \sum_{s \in S \setminus O} \sum_{x_O \in \mathcal{X}_O} p_O(x_O) f_s(z_I, z_C, x_O)).$$

Hence

$$h := \sum_{s \in S \setminus O} \sum_{x_O \in \mathcal{X}_O} p_O(x_O) f_s(\cdot, \cdot, x_O) \leq 0.$$

De Cooman & MirandaThe gambles $f_s(\cdot, \cdot, x_O)$ on $\mathcal{X}_{I \cup C}$ [where $x_O \in \mathcal{X}_O$ and $s \in S \setminus O$] can clearly not all be zero. The non-zero ones all belong to $\otimes_{s \in I \cup C} \mathcal{D}_s$, by Lemma 16. But then the coherence of the set of desirable gambles $\otimes_{s \in I \cup C} \mathcal{D}_s$ [Proposition 15] guarantees that their positive linear combination $h$ is an element of $\otimes_{c \in C} \mathcal{D}_c$ for which $h \leq 0$, a contradiction. Hence indeed $f \in \otimes_{o \in O} \mathcal{D}_o$. $\square$

**Theorem 19 (Independent natural extension).** *Consider the coherent sets $\mathcal{D}_n$ of desirable gambles on $\mathcal{X}_n$, $n \in N$. Then $\otimes_{n \in N} \mathcal{D}_n$ is the smallest coherent set of desirable gambles on $\mathcal{X}_N$ that is an independent product of the coherent sets of desirable gambles $\mathcal{D}_n$, $n \in N$.*

We call $\otimes_{n \in N} \mathcal{D}_n$ the *independent natural extension* of the marginals $\mathcal{D}_n$.

*Proof.* It follows from Propositions 15, 17 and 18 that $\otimes_{n \in N} \mathcal{D}_n$ is an independent product $\mathcal{D}_N$ of the $\mathcal{D}_n$. To prove that it is the smallest one, consider any independent product $\mathcal{D}_N$ of the $\mathcal{D}_n$. Fix $n \in N$. If we consider any $x_{N \setminus \{n\}} \in \mathcal{X}_{N \setminus \{n\}}$, then $\mathrm{marg}_n(\mathcal{D}_N \rfloor x_{N \setminus \{n\}}) = \mathcal{D}_n$, by assumption. If we therefore consider any $g \in \mathcal{D}_n$, this in turn implies that $g \in \mathcal{D}_N \rfloor x_{N \setminus \{n\}}$, and therefore $\mathbb{I}_{\{x_{N \setminus \{n\}}\}} g \in \mathcal{D}_N$, by Eq. (9). So we infer by coherence that $\mathcal{A}^{\mathrm{irr}}_{N \setminus \{n\} \to \{n\}} \subseteq \mathcal{D}_N$, and therefore also that $\otimes_{n \in N} \mathcal{D}_n \subseteq \mathcal{D}_N$. $\square$

One of the most useful properties of the independent natural extension, is its associativity: it allows us to construct the extension in a modular fashion.

**Theorem 20 (Associativity of independent natural extension).** *Let $N_1$ and $N_2$ be disjoint non-empty index sets, and consider $\mathcal{D}_{n_k} \in \mathbb{D}(\mathcal{X}_{n_k})$, $n_k \in N_k$, $k = 1, 2$. Then given $\mathcal{D}_{N_1} \coloneqq \otimes_{n_1 \in N_1} \mathcal{D}_{n_1}$ and $\mathcal{D}_{N_2} \coloneqq \otimes_{n_2 \in N_2} \mathcal{D}_{n_2}$, it holds that*

$$\mathcal{D}_{N_1} \otimes \mathcal{D}_{N_2} = \otimes_{n \in N_1 \cup N_2} \mathcal{D}_n.$$

*Proof.* We first prove that $\mathcal{D}_{N_1} \otimes \mathcal{D}_{N_2} \subseteq \otimes_{n \in N_1 \cup N_2} \mathcal{D}_n$. Fix any gamble $h \in \mathcal{A}^{\mathrm{irr}}_{\{N_1\} \to \{N_2\}}$ and any $x_{N_1} \in \mathcal{X}_{N_1}$, so $h(x_{N_1}, \cdot) \in \mathcal{D}_{N_2} \cup \{0\}$ by Eq. (17). It follows from Eq. (18) that there are gambles $h^{n_2}_{x_{N_1}} \in \mathcal{A}^{\mathrm{irr}}_{N_2 \setminus \{n_2\} \to \{n_2\}} \cup \{0\}$ for all $n_2 \in N_2$ such that

$$h(x_{N_1}, \cdot) \geq \sum_{n_2 \in N_2} h^{n_2}_{x_{N_1}}.$$

Define, for any $n_2 \in N_2$, the gamble $g_{n_2}$ on $\mathcal{X}_N$ by letting $g_{n_2}(x_{N \setminus \{n_2\}}, \cdot) \coloneqq h^{n_2}_{x_{N_1}}(x_{N_2 \setminus \{n_2\}}, \cdot)$ for all $x_N \in \mathcal{X}_N$. Then it follows from Eq. (17) that $g_{n_2}(x_{N \setminus \{n_2\}}, \cdot) \in \mathcal{D}_{n_2} \cup \{0\}$ for all $x_N \in \mathcal{X}_N$, and therefore $g_{n_2} \in \mathcal{A}^{\mathrm{irr}}_{N \setminus \{n_2\} \to \{n_2\}} \cup \{0\}$. Moreover,

$$h = \sum_{x_{N_1} \in \mathcal{X}_{N_1}} \mathbb{I}_{\{x_{N_1}\}} h(x_{N_1}, \cdot)$$

$$\geq \sum_{x_{N_1} \in \mathcal{X}_{N_1}} \mathbb{I}_{\{x_{N_1}\}} \sum_{n_2 \in N_2} h^{n_2}_{x_{N_1}} = \sum_{n_2 \in N_2} \sum_{x_{N_1} \in \mathcal{X}_{N_1}} \mathbb{I}_{\{x_{N_1}\}} h^{n_2}_{x_{N_1}} = \sum_{n_2 \in N_2} g_{n_2},$$

It therefore follows from Eq. (18) that $h \in \otimes_{n \in N_1 \cup N_2} \mathcal{D}_n$, since clearly $h \neq 0$ because of Eq. (17). We conclude that $\mathcal{A}^{\mathrm{irr}}_{\{N_1\} \to \{N_2\}} \subseteq \otimes_{n \in N_1 \cup N_2} \mathcal{D}_n$. Similarly, we can prove the inclusion $\mathcal{A}^{\mathrm{irr}}_{\{N_2\}\to\{N_1\}} \subseteq \otimes_{n\in N_1\cup N_2}\mathcal{D}_n$, and therefore also $\mathcal{D}_{N_1} \otimes \mathcal{D}_{N_2} \subseteq \otimes_{n\in N_1\cup N_2}\mathcal{D}_n$, again by Eq. (18).

Next, we prove the converse inclusion $\otimes_{n\in N_1\cup N_2}\mathcal{D}_n \subseteq \mathcal{D}_{N_1} \otimes \mathcal{D}_{N_2}$. Consider any gamble $h \in \otimes_{n\in N_1\cup N_2}\mathcal{D}_n$, then by Eq. (18) there are $h_n \in \mathcal{A}^{\mathrm{irr}}_{N_1\cup N_2\setminus\{n\}\to\{n\}} \cup \{0\}$ for all $n \in N_1 \cup N_2$ such that

$$h \geq \sum_{n\in N} h_n = h_1 + h_2, \text{ where } h_1 := \sum_{n_1\in N_1} h_{n_1} \text{ and } h_2 := \sum_{n_2\in N_2} h_{n_2}.$$

Fix any $x_{N_1} \in \mathcal{X}_{N_1}$. For any $n_2 \in N_2$, $h_{n_2} \in \mathcal{A}^{\mathrm{irr}}_{N_1\cup N_2\setminus\{n_2\}\to\{n_2\}} \cup \{0\}$ implies that $h_{n_2}(x_{N_1},\cdot) \in \mathcal{A}^{\mathrm{irr}}_{N_2\setminus\{n_2\}\to\{n_2\}} \cup \{0\}$ by Lemma 16. Hence $h_2(x_{N_1},\cdot) \in \mathcal{D}_{N_2} \cup \{0\}$ by Eq. (18), and therefore $h_2 \in \mathcal{A}^{\mathrm{irr}}_{\{N_1\}\to\{N_2\}} \cup \{0\}$ by Eq. (17). Similarly, $h_1 \in \mathcal{A}^{\mathrm{irr}}_{\{N_2\}\to\{N_1\}} \cup \{0\}$, and therefore $h \in \mathcal{D}_{N_1} \otimes \mathcal{D}_{N_2}$ by Eq. (18), since clearly $h \neq 0$. □

To conclude this section, we establish a connection between independent natural extension for sets of desirable gambles and the eponymous notion for coherent lower previsions, studied in detail by De Cooman et al. (2011). Given coherent lower previsions $\underline{P}_n$ on $\mathcal{G}(\mathcal{X}_n)$, $n \in N$, their *independent natural extension* is the coherent lower prevision given by

$$\underline{E}_N(f) := \sup_{\substack{h_n\in\mathcal{G}(\mathcal{X}_N) \\ n\in N}} \min_{z_N\in\mathcal{X}_N} \left[ f(z_N) - \sum_{n\in N}[h_n(z_N) - \underline{P}_n(h_n(\cdot, z_{N\setminus\{n\}}))]\right] \quad (19)$$

for all gambles $f$ on $\mathcal{X}_N$. It is the point-wise smallest (most conservative) joint lower prevision that satisfies the property of coherence by Walley (1991, ch. 7) with the marginals $\underline{P}_n$ given an assessment of epistemic independence of the variables $X_n$, $n \in N$.

The correspondence between coherent lower previsions and sets of desirable gambles has been mentioned in Section 2.6; we show next that if we have such a correspondence between the marginals, it also holds between their associated independent natural extensions.

**Theorem 21.** *Let $\mathcal{D}_n$ be coherent sets of desirable gambles on $\mathcal{X}_n$ for $n \in N$, and let $\otimes_{n\in N}\mathcal{D}_n$ be their independent natural extension. Consider the coherent lower previsions $\underline{P}_n$ on $\mathcal{G}(\mathcal{X}_n)$ given by $\underline{P}_n(f_n) := \sup\{\mu \in \mathbb{R}: f_n - \mu \in \mathcal{D}_n\}$ for all $f_n \in \mathcal{G}(\mathcal{X}_n)$. Then the independent natural extension $\underline{E}_N$ of the marginal lower previsions $\underline{P}_n$, $n \in N$ satisfies*

$$\underline{E}_N(f) = \sup\{\mu \in \mathbb{R}: f - \mu \in \otimes_{n\in N}\mathcal{D}_n\} \text{ for all } f \in \mathcal{G}(\mathcal{X}_N).$$

*Proof.* Fix any gamble $f$ in $\mathcal{G}(\mathcal{X}_N)$. First, consider any real number $\mu < \underline{E}_N(f)$, then it follows from Eq. (19) that there are $\delta > 0$ and $h_n \in \mathcal{G}(\mathcal{X}_N)$, $n \in N$ such that $f - \mu \geq \sum_{n\in N} g_n$, where we defined the gambles $g_n$ on $\mathcal{X}_N$ by $g_n(z_N) := h_n(z_N) - \underline{P}_n(h_n(z_{N\setminus\{n\}},\cdot)) + \delta$ for all $z_N \in \mathcal{X}_N$. But it follows from the definition of $\underline{P}_n$ that

$$g_n(z_{N\setminus\{n\}},\cdot) = h_n(z_{N\setminus\{n\}},\cdot) - \underline{P}_n(h_n(z_{N\setminus\{n\}},\cdot)) + \delta \in \mathcal{D}_n \text{ for all } z_{N\setminus\{n\}} \in \mathcal{X}_{N\setminus\{n\}}.$$

Since clearly $g_n \neq 0$, Eq. (17) then tells us that $g_n \in \mathcal{A}^{\mathrm{irr}}_{N\setminus\{n\}\to\{n\}}$, and we infer from Eq. (18) that $\sum_{n\in N} g_n \in \otimes_{n\in N}\mathcal{D}_n$, and therefore also $f - \mu \in \otimes_{n\in N}\mathcal{D}_n$. This guarantees that $\underline{E}_N(f) \leq \sup\{\mu \in \mathbb{R}: f - \mu \in \otimes_{n\in N}\mathcal{D}_n\}$.

To prove the converse inequality, consider any real number $\mu$ such that $f - \mu \in \otimes_{n \in N} \mathcal{D}_n$. We infer using Eq. (18) that there are gambles $h_n \in \mathcal{A}^{\text{irr}}_{N \setminus \{n\} \to \{n\}}$, $n \in N$ such that $f - \mu \geq \sum_{n \in N} h_n$. For all $n \in N$ and $z_{N \setminus \{n\}} \in \mathcal{X}_{N \setminus \{n\}}$, it follows from Eq. (17) that $h_n(z_{N \setminus \{n\}}, \cdot) \in \mathcal{D}_n \cup \{0\}$, and therefore $\underline{P}_n(h_n(z_{N \setminus \{n\}}, \cdot)) \geq 0$, whence

$$\sum_{n \in N} \left[ h_n(z_N) - \underline{P}_n(h_n(z_{N \setminus \{n\}}, \cdot)) \right] \leq \sum_{n \in N} h_n(z_N) \leq f(z_N) - \mu.$$

We then infer from Eq. (19) that $\underline{E}_N(f) \geq \mu$ and so we find that indeed also $\underline{E}_N(f) \geq \sup \{\mu \in \mathbb{R} : f - \mu \in \otimes_{n \in N} \mathcal{D}_n\}$. □

In a similar way as for the irrelevant natural extension, we infer from Eqs. (16) and (18) that the computational complexity of finding or representing the independent natural extension of a number of marginal models $\mathcal{D}_n$ is linear in the number of extreme rays of the $\mathcal{D}_n$, and linear in the number of elements of the sets $\mathcal{X}_{N \setminus \{n\}}$—and therefore essentially exponential in the number $|N|$ of independent variables $X_n$, $n \in N$. Similar results will hold in the more general case that the marginal sets of desirable gambles can be characterised using a finite number of 'generalised' extreme rays, as described by Couso and Moral (2011) and Quaeghebeur (2012a).

## 8. Maximal Coherent Sets of Desirable Gambles and Strong Products

We have seen that for any collection $\mathcal{D}_n$, $n \in N$ of marginal coherent sets of desirable gambles, there always is a *smallest* independent product, which we have called the independent natural extension $\otimes_{n \in N} \mathcal{D}_n$. We have proceeded in this way because we had no way of excluding that there may be other, larger, independent products. Indeed, we show in this section that such is the case. Using the notions of independent natural extension and maximal coherent sets of desirable gambles, we can consistently define a specific independent product that typically strictly includes the independent natural extension. We call it the *strong product*, because it is very close in spirit to the strong product used in coherent lower prevision theory (Couso, Moral, & Walley, 2000; Cozman, 2000, 2005; De Cooman et al., 2011), as we shall see in Theorem 28.

### 8.1 Independent Products of Maximal Coherent Sets of Desirable Gambles

We begin by mentioning a number of interesting facts about maximal coherent sets of desirable gambles, and their independent products. The following result was already (essentially) proved by Couso and Moral (2011): updating a coherent set of desirable gambles preserves its maximality.

**Proposition 22.** *Let $\mathcal{M}_N \in \mathbb{M}(\mathcal{X}_N)$, and consider any disjoint subsets $I$ and $O$ of $N$. Then $\text{marg}_O(\mathcal{M}_N \rfloor x_I) \in \mathbb{M}(\mathcal{X}_O)$ for all $x_I \in \mathcal{X}_I$.*

*Proof.* Suppose there is some $x_I$ in $\mathcal{X}_I$ for which $\text{marg}_O(\mathcal{M}_N \rfloor x_I)$ is not maximal. This means that there is some $f \in \mathcal{G}(\mathcal{X}_O)$ for which neither $f$ nor $-f$ belong to $\mathcal{M}_N \rfloor x_I$, which in turn implies that neither $\mathbb{I}_{\{x_I\}} f$ nor $-\mathbb{I}_{\{x_I\}} f$ belong to $\mathcal{M}_N$. But this contradicts the maximality of $\mathcal{M}_N$. □

On the other hand, taking the independent natural extension does not necessarily preserve maximality: if $\mathcal{M}_n \in \mathbb{M}(\mathcal{X}_n)$ for all $n \in N$, then it does not necessarily hold that $\otimes_{n \in N} \mathcal{M}_n \in \mathbb{M}(\mathcal{X}_N)$, as the counterexample in Section A.1 shows. Interestingly, that example does not present an isolated case: when we consider two binary variables, the independent natural extension of two maximal coherent sets of desirable gambles is *never* maximal, as we can see in our next proposition. It is an open problem whether this negative result can be extended to any finite set of (not necessarily binary) variables.

An intuitive explanation of this result is that each of the maximal sets of gambles is a half-space where we are excluding one of the two rays determining its boundary, so as not to have the zero gamble as desirable; and when we apply the notion of independent natural extension we end up missing three of the four parts of the boundary of the set of gambles in the product space, preventing this product from being maximal.

**Proposition 23.** *Consider $\mathcal{X}_1 = \mathcal{X}_2 = \{0, 1\}$, and let $\mathcal{M}_1$ and $\mathcal{M}_2$ be any maximal coherent sets of desirable gambles on $\mathcal{X}_1$ and $\mathcal{X}_2$, respectively. Then their independent natural extension $\mathcal{M}_1 \otimes \mathcal{M}_2$ is not a maximal coherent set of desirable gambles.*

*Proof.* Let $p_k$ be the mass function of the linear prevision $P_k$ determined by $\mathcal{M}_k$, $k = 1, 2$. We deduce from Theorem 21 that the lower prevision determined by $\mathcal{M}_1 \otimes \mathcal{M}_2$ is the independent natural extension of the linear previsions $P_1$ and $P_2$, and therefore equal to the independent product $P_{\{1,2\}}$ of these linear previsions [see Proposition 25 in De Cooman et al., 2011]. This is the linear prevision on $\mathcal{G}(\mathcal{X}_{\{1,2\}})$ with mass function defined by $p_{\{1,2\}}(x_1, x_2) := p_1(x_1)p_2(x_2)$ for all $(x_1, x_2) \in \mathcal{X}_{\{1,2\}}$.

Before we really get the proof on the tracks, we make a useful observation. Any maximal $\mathcal{M}_k$ is a semi-plane through the origin that excludes the origin, includes its boundary on one side of the origin, and excludes the boundary on the other side. This means that there is unique element $a_k$ of $\mathcal{X}_k$ where all the elements $f_k$ of the included boundary—those elements $f_k$ of $\mathcal{M}_k$ for which $P_k(f_k)$ is zero—are positive $f_k(a_k) > 0$. We denote the single other element of $\mathcal{X}_k$ by $b_k$. In other words, if we express

$$\mathcal{M}_k = \{f_k : P_k(f_k) > 0\} \cup \{f_k : P_k(f_k) = 0, f_k \in \mathcal{M}_k\},$$

and consider $f_k \in \mathcal{M}_k$ with $P_k(f_k) = p_k(a_k)f_k(a_k) + p_k(b_k)f_k(b_k) = 0$, then if $f_k(a_k) > 0$ there cannot be any $g_k \in \mathcal{M}_k$ with $P_k(g_k) = 0$ and $g_k(b_k) > 0$: otherwise, the zero gamble would be a convex combination of $f_k$ and $g_k$ [it would be $0 = f_k - \frac{f_k(b_k)}{g_k(b_k)} g_k$] and it would thus belong to $\mathcal{M}_k$, a contradiction with its coherence. Note that in this reasoning we assume implicitly that $p_k(a_k) \in (0,1)$; otherwise, if for instance $p_k(a_k) = 0$, a gamble $f_k$ satisfies $P_k(f_k) = 0$ if and only if $f_k(b_k) = 0$, and then $f_k$ can only belong to $\mathcal{M}_k$ if $f_k(a_k) > 0$.

We are now ready to turn to the proof. There are a number of possibilities.

First, assume that both $p_k(a_k) > 0$ and $p_k(b_k) > 0$ for $k = 1, 2$. Consider any gamble $h$ on $\mathcal{X}_{\{1,2\}}$ such that $h(a_1, a_2) = h(b_1, b_2) = 0$, $\min h < 0$, $\max h > 0$ and

$$P_{\{1,2\}}(h) = p_1(a_1)p_2(b_2)h(a_1, b_2) + p_1(b_1)p_2(a_2)h(b_1, a_2) = 0.$$

Of course, there always is such a gamble, and we are going to show that it does not belong to $\mathcal{M}_1 \otimes \mathcal{M}_2$.

Assume *ex absurdo* that it does, meaning that there are $h_1 \in \mathcal{A}^{\text{irr}}_{\{2\}\to\{1\}}$ and $h_2 \in \mathcal{A}^{\text{irr}}_{\{1\}\to\{2\}}$ such that $h \geq h_1 + h_2$. By definition, $h_1(\cdot, x_2) \in \mathcal{M}_1 \cup \{0\}$ and therefore $P_1(h_1(\cdot, x_2)) \geq 0$ for all $x_2 \in \mathcal{X}_2$. Similarly, $P_2(h_2(x_1, \cdot)) \geq 0$ for all $x_1 \in \mathcal{X}_1$. Hence $0 = P_{\{1,2\}}(h) \geq P_{\{1,2\}}(h_1) + P_{\{1,2\}}(h_2) \geq 0$, taking into account that

$$P_{\{1,2\}}(h_1) = \sum_{x_2 \in \mathcal{X}_2} p_2(x_2) P_1(h_1(\cdot, x_2)) \geq 0 \text{ and } P_{\{1,2\}}(h_2) = \sum_{x_1 \in \mathcal{X}_1} p_1(x_1) P_2(h_2(x_1, \cdot)) \geq 0.$$

As a consequence, $P_{\{1,2\}}(h_1) = P_{\{1,2\}}(h_2) = 0$. But this in turn implies that $P_1(h_1(\cdot, x_2)) = 0$ for all $x_2 \in \mathcal{X}_2$ and that $P_2(h_2(x_1, \cdot)) = 0$ for all $x_1 \in \mathcal{X}_1$. Given the observations made at the start of the proof, we therefore come to the conclusion that $h_1(a_1, x_2) \geq 0$ for all $x_2 \in \mathcal{X}_2$ and $h_2(x_1, a_2) \geq 0$ for all $x_1 \in \mathcal{X}_1$. But then $h(a_1, a_2) = 0$ implies that $h_1(a_1, a_2) = h_2(a_1, a_2) = 0$, which in turn implies that $h_1(b_1, a_2) = h_2(a_1, b_2) = 0$, because both $0 = P_1(h_1(\cdot, a_2)) = p_1(a_1) h_1(a_1, a_2) + p_1(b_1) h_1(b_1, a_2)$ and $0 = P_2(h_2(a_1, \cdot)) = p_2(a_1) h_1(a_1, a_2) + p_2(b_1) h_1(a_1, b_2)$. So we eventually find that

$$h(b_1, a_2) \geq h_1(b_1, a_2) + h_2(b_1, a_2) \geq 0 \text{ and } h(a_1, b_2) \geq h_1(a_1, b_2) + h_2(a_1, b_2) \geq 0,$$

which contradicts $\min h < 0$.

Now, if any non-zero $h$ such that $h(a_1, a_2) = h(b_1, b_2) = 0 = P_{\{1,2\}}(h)$ with $\min h < 0$ and $\max h > 0$ does not belong to $\mathcal{M}_1 \otimes \mathcal{M}_2$, neither does $-h$, and this means that $\mathcal{M}_1 \otimes \mathcal{M}_2$ is not maximal.

Next we consider the cases where one of the marginal linear previsions are degenerate. Assume for instance that $p_1(a_1) = 0$ and $p_2(a_2) \in (0, 1)$ [the other cases where only one of the marginals is degenerate are similar]. Consider a non-zero gamble $h_2 \notin \mathcal{M}_2$ such that $P_2(h_2) = 0$ [always possible]. Then $-h_2 \in \mathcal{M}_2$ and it follows from the observations made in the beginning of this proof that $h_2(a_2) < 0$. Now consider the gamble $h$ defined by

$$h(b_1, a_2) := h_2(a_2) < 0, \quad h(b_1, b_2) := h_2(b_2) \geq 0, \quad h(a_1, a_2) = h(a_1, a_2) := 1.$$

It follows that $P_{\{1,2\}}(h) = P_2(h_2) = 0$. To see that $h \notin \mathcal{M}_1 \otimes \mathcal{M}_2$, assume that there are $f_1 \in \mathcal{A}^{\text{irr}}_{\{2\}\to\{1\}}$ and $f_2 \in \mathcal{A}^{\text{irr}}_{\{1\}\to\{2\}}$ such that $h \geq f_1 + f_2$. But then $0 = P_{\{1,2\}}(h) \geq P_{\{1,2\}}(f_1) + P_{\{1,2\}}(f_2) \geq 0$ and therefore $0 = P_{\{1,2\}}(f_1) = p_2(a_2) f_1(b_1, a_2) + p_2(b_2) f_1(b_1, b_2)$. On the other hand, $f_1 \in \mathcal{A}^{\text{irr}}_{\{2\}\to\{1\}}$ also implies that $P_1(f_1(\cdot, x_2)) \geq 0$ for all $x_2 \in \mathcal{X}_2$, and therefore $f_1(b_1, a_2) \geq 0$ and $f_1(b_1, b_2) \geq 0$. Hence $f_1(b_1, \cdot) = 0$, and therefore $f_2(b_1, \cdot) \leq h(b_1, \cdot) = h_2$ and since $h_2 \notin \mathcal{M}_2$, it follows that $f_2(b_1, \cdot) \notin \mathcal{M}_2$. Because we must have by definition that $f_2(b_1, \cdot) \in \mathcal{M}_2 \cup \{0\}$, this can only mean that $f_2(b_1, \cdot) = 0$, whence $h_2 \geq 0$, contradicting $h_2(a_2) < 0$. This implies that $h$ cannot belong to $\mathcal{M}_1 \otimes \mathcal{M}_2$.

Similarly, if $-h$ belongs to $\mathcal{M}_1 \otimes \mathcal{M}_2$, then there must be $g_1 \in \mathcal{A}^{\text{irr}}_{\{2\}\to\{1\}}$ and $g_2 \in \mathcal{A}^{\text{irr}}_{\{1\}\to\{2\}}$ such that $-h \geq g_1 + g_2$. But then $0 = P_{\{1,2\}}(-h) \geq P_{\{1,2\}}(g_1) + P_{\{1,2\}}(g_2) \geq 0$, whence $0 = P_{\{1,2\}}(g_1) = p_2(a_2) g_1(b_1, a_2) + p_2(b_2) g_1(b_1, b_2)$. But $g_1 \in \mathcal{A}^{\text{irr}}_{\{2\}\to\{1\}}$ also implies that $P_1(g_1(\cdot, x_2)) \geq 0$ for all $x_2 \in \mathcal{X}_2$, and therefore $g_1(b_1, a_2) \geq 0$ and $g_1(b_1, b_2) \geq 0$. Hence $g_1(b_1, \cdot) = 0$, and therefore we find that $g_1(a_1, a_2) \geq 0$ and $g_1(a_1, b_2) \geq 0$ [if, say, $g_1(a_1, a_2) < 0$ then $g_1(\cdot, a_2) < 0$ because also $g_1(b_1, a_2) = 0$, which contradicts that $g_1(\cdot, a_2) \in \mathcal{M}_1 \cup \{0\}$, a consequence of $g_1 \in \mathcal{A}^{\text{irr}}_{\{2\}\to\{1\}}$]. Since, moreover, $g_2 \in \mathcal{A}^{\text{irr}}_{\{1\}\to\{2\}}$ implies that $0 \leq P_2(g_2(a_1, \cdot)) = p_2(a_2) g_2(a_1, a_2) + p_2(b_2) g_2(a_1, b_2)$ and therefore also that $g_2(a_1, a_2) \geq 0$ or $g_2(a_1, b_2) \geq 0$, it follows that $-h(a_1, a_2) \geq 0$ or $-h(a_1, b_2) \geq 0$, which contradicts $-h(a_1, a_2) = -h(a_1, b_2) = -1 < 0$. Hence, $-h$ does not belong to $\mathcal{M}_1 \otimes \mathcal{M}_2$ either, so $\mathcal{M}_1 \otimes \mathcal{M}_2$ is not maximal.

Finally, we turn to the cases where all marginals are degenerate. Assume for instance that $p_1(a_1) = p_2(a_2) = 0$ [the other cases where both marginals are degenerate, are similar]. Consider the gamble $h$ given by

$$h(a_1, a_2) = h(b_1, b_2) = 0, \quad h(b_1, a_2) = 1, \quad h(a_1, b_2) = -1,$$

then $P_{\{1,2\}}(h) = p_1(b_1)p_2(b_2)h(b_1, b_2) = 0$. To see that $h \notin \mathcal{M}_1 \otimes \mathcal{M}_2$, assume *ex absurdo* that there are $u_1 \in \mathcal{A}^{\text{irr}}_{\{2\} \to \{1\}}$ and $u_2 \in \mathcal{A}^{\text{irr}}_{\{1\} \to \{2\}}$ such that $h \geq u_1 + u_2$. Then $u_1 \in \mathcal{A}^{\text{irr}}_{\{2\} \to \{1\}}$ implies that

$$u_1(b_1, b_2) = P_1(u_1(\cdot, b_2)) \geq 0 \text{ and } u_1(b_1, a_2) = P_1(u_1(\cdot, a_2)) \geq 0,$$

and similarly $u_2 \in \mathcal{A}^{\text{irr}}_{\{1\} \to \{2\}}$ implies that $u_2(b_1, b_2) = P_2(u_2(b_1, \cdot)) \geq 0$ and $u_2(a_1, b_2) = P_2(u_2(a_1, \cdot)) \geq 0$. Now it also follows from $P_{\{1,2\}}(h) = 0$, $P_{\{1,2\}}(u_1) \geq 0$ and $P_{\{1,2\}}(u_2) \geq 0$ that $u_1(b_1, b_2) = P_{\{1,2\}}(u_1) = 0$ and $u_2(b_1, b_2) = P_{\{1,2\}}(u_2) = 0$, and as a consequence we find that $u_1(a_1, b_2) \geq 0$ and $u_2(b_1, a_2) \geq 0$ [if, say, $u_1(a_1, b_2) < 0$ then $u_1(\cdot, b_2) < 0$ because also $u_1(b_1, b_2) = 0$, which contradicts $u_1(\cdot, b_2) \in \mathcal{M}_1 \cup \{0\}$, a consequence of $u_1 \in \mathcal{A}^{\text{irr}}_{\{2\} \to \{1\}}$]. As a consequence, $-1 = h(a_1, b_2) \geq u_1(a_1, b_2) + u_2(a_1, b_2) \geq 0$, a contradiction. Hence indeed, $h$ does not belong to $\mathcal{M}_1 \otimes \mathcal{M}_2$.

Finally, assume *ex absurdo* that there are $v_1 \in \mathcal{A}^{\text{irr}}_{\{2\} \to \{1\}}$ and $v_2 \in \mathcal{A}^{\text{irr}}_{\{1\} \to \{2\}}$ such that $-h \geq v_1 + v_2$. Then $v_1 \in \mathcal{A}^{\text{irr}}_{\{2\} \to \{1\}}$ implies that

$$v_1(b_1, b_2) = P_1(v_1(\cdot, b_2)) \geq 0 \text{ and } v_1(b_1, a_2) = P_1(v_1(\cdot, a_2)) \geq 0,$$

and similarly $v_2 \in \mathcal{A}^{\text{irr}}_{\{1\} \to \{2\}}$ implies that $v_2(b_1, b_2) = P_2(v_2(b_1, \cdot)) \geq 0$ and $v_2(a_1, b_2) = P_2(v_2(a_1, \cdot)) \geq 0$. Now it also follows from $P_{\{1,2\}}(-h) = 0$, $P_{\{1,2\}}(v_1) \geq 0$ and $P_{\{1,2\}}(v_1) \geq 0$ that $v_1(b_1, b_2) = P_{\{1,2\}}(v_1) = 0$ and $v_2(b_1, b_2) = P_{\{1,2\}}(v_2) = 0$, and as a consequence we find that $v_1(a_1, b_2) \geq 0$ and $v_2(b_1, a_2) \geq 0$ [if, say, $v_1(a_1, b_2) < 0$ then $v_1(\cdot, b_2) < 0$ because also $v_1(b_1, b_2) = 0$, which contradicts $v_1(\cdot, b_2) \in \mathcal{M}_1 \cup \{0\}$, a consequence of $v_1 \in \mathcal{A}^{\text{irr}}_{\{2\} \to \{1\}}$]. As a consequence, $-1 = -h(b_1, a_2) \geq v_1(b_1, a_2) + v_2(b_1, a_2) \geq 0$, a contradiction. This shows that $-h$ does not belong to $\mathcal{M}_1 \otimes \mathcal{M}_2$ either, and therefore this set is not maximal. □

On the other hand, the Example A.2 in the Appendix shows that there are independent products of maximal coherent sets of desirable gambles that are maximal; hence, the independent natural extension of maximal coherent sets is not their only independent product. Indeed, we can establish the following result:

**Proposition 24.** *Consider maximal coherent sets of desirable gambles $\mathcal{M}_1 \in \mathbb{M}(\mathcal{X}_1)$ and $\mathcal{M}_2 \in \mathbb{M}(\mathcal{X}_2)$.*

(i) *Let $\mathcal{D}_{\{1,2\}}$ be any coherent set of desirable gambles on $\mathcal{X}_{\{1,2\}}$ such that $\mathcal{M}_1 \otimes \mathcal{M}_2 \subseteq \mathcal{D}_{\{1,2\}}$. Then $\mathcal{D}_{\{1,2\}}$ is independent with marginals $\mathcal{M}_1$ and $\mathcal{M}_2$.*

(ii) *As a consequence, a maximal set of gambles $\mathcal{M}_{\{1,2\}}$ is an independent product of its marginals if and only if $\mathcal{M}_{\{1,2\}} \rfloor x_2$ is the same for all $x_2 \in \mathcal{X}_2$ and $\mathcal{M}_{\{1,2\}} \rfloor x_1$ is the same for all $x_1 \in \mathcal{X}_1$.*

*Proof.* (i). We have for every $x_1 \in \mathcal{X}_1$ that $\mathcal{M}_2 = (\mathcal{M}_1 \otimes \mathcal{M}_2)\rfloor x_1 \subseteq \mathcal{D}_{\{1,2\}}\rfloor x_1$, where the equality follows from Proposition 18. Since $\mathcal{M}_2$ is maximal, this implies that $\mathcal{M}_2 = \mathcal{D}_{\{1,2\}}\rfloor x_1$ for all $x_1 \in \mathcal{X}_1$, and a similar argument shows that $\mathcal{M}_1 = \mathcal{D}_{\{1,2\}}\rfloor x_2$ for all $x_2 \in \mathcal{X}_2$. On the other hand, it follows from Proposition 17 that $\mathcal{M}_2 = \mathrm{marg}_2(\mathcal{M}_1 \otimes \mathcal{M}_2) \subseteq \mathrm{marg}_2(\mathcal{D}_{\{1,2\}})$. Since $\mathcal{M}_2$ is maximal, this implies that $\mathcal{M}_2 = \mathrm{marg}_2(\mathcal{D}_{\{1,2\}})$, and a similar argument shows that $\mathcal{M}_1 = \mathrm{marg}_1(\mathcal{D}_{\{1,2\}})$. In summary, we see that $\mathrm{marg}_1(\mathcal{D}_{\{1,2\}}) = \mathcal{D}_{\{1,2\}}\rfloor x_2$ for all $x_2 \in \mathcal{X}_2$, and $\mathrm{marg}_2(\mathcal{D}_{\{1,2\}}) = \mathcal{D}_{\{1,2\}}\rfloor x_1$ for all $x_1 \in \mathcal{X}_1$, showing that $\mathcal{D}_{\{1,2\}}$ is indeed independent.

(ii). It follows from the definition of an independent product that it is necessary that $\mathcal{M}_{\{1,2\}}\rfloor x_2$ and $\mathcal{M}_{\{1,2\}}\rfloor x_1$ should be the same for all $x_2$ and $x_1$, respectively. To see that this is also a sufficient condition for $\mathcal{M}_{\{1,2\}}$ to be an independent product, note that in that case $\mathcal{M}_{\{1,2\}}\rfloor x_1 \otimes \mathcal{M}_{\{1,2\}}\rfloor x_2 \subseteq \mathcal{M}_{\{1,2\}}$, and that the sets $\mathcal{M}_{\{1,2\}}\rfloor x_1$ and $\mathcal{M}_{\{1,2\}}\rfloor x_2$ are maximal, by Proposition 22. On the other hand, Proposition 17 implies that

$$\mathrm{marg}_1(\mathcal{M}_{\{1,2\}}\rfloor x_1 \otimes \mathcal{M}_{\{1,2\}}\rfloor x_2) = \mathcal{M}_{\{1,2\}}\rfloor x_1 \subseteq \mathrm{marg}_1(\mathcal{M}_{\{1,2\}}),$$

so both sets are equal. Similarly, we deduce that

$$\mathrm{marg}_2(\mathcal{M}_{\{1,2\}}\rfloor x_1 \otimes \mathcal{M}_{\{1,2\}}\rfloor x_2) = \mathcal{M}_{\{1,2\}}\rfloor x_2 \subseteq \mathrm{marg}_2(\mathcal{M}_{\{1,2\}}),$$

and therefore $\mathrm{marg}_1(\mathcal{M}_{\{1,2\}}) \otimes \mathrm{marg}_2(\mathcal{M}_{\{1,2\}}) \subseteq \mathcal{M}_{\{1,2\}}$. Invoking the first part of the proposition, we find that $\mathcal{M}_{\{1,2\}}$ is an independent product of its marginals. $\square$

The first part of this proposition provides us with a simple characterisation of the independent products of two maximal sets: they are simply those coherent supersets of the independent natural extension; in particular, this means that any maximal superset of this independent natural extension will be an independent product, so two maximal sets always have maximal products (although these will differ from the independent natural extension). The second part implies that if the sets of conditional gambles coincide for all the conditioning events, then they automatically agree with the marginal sets of gambles, and as a consequence the set is an independent product.

### 8.2 The Strong Product and Its Properties

Now consider the case where we have coherent marginal sets of desirable gambles $\mathcal{D}_n$ for all $n \in N$. We define their *strong product* $\boxtimes_{n \in N} \mathcal{D}_n$ as the set of desirable gambles on the product space $\mathcal{X}_N$ given by:[8]

$$\boxtimes_{n \in N} \mathcal{D}_n := \bigcap \{\otimes_{n \in N} \mathcal{M}_n \colon \mathcal{M}_n \in m(\mathcal{D}_n), n \in N\},$$

where $m(\mathcal{D}_n)$ is given by Eq. (1). This strong product corresponds is the set of desirable gambles determined by a notion of independence that is more restrictive than those of epistemic irrelevance and independence considered so far: that of *strong independence* (Couso et al., 2000; Cozman, 2012), sometimes called *type-3 independence* (de Campos & Moral, 1995). Strong independence means that the associated joint credal set is the convex hull of the set of linear previsions that are the stochastic independent products of linear previsions that dominate the marginals; or, equivalently, that the associated lower prevision is the lower envelope of the products of the linear previsions that dominate the marginals. This will be clearer after Theorem 28.

---

8. As this paper focusses on independent natural extension, because that has a much more direct behavioural justification, we will forgo discussing the complexity of computing this strong product, which, on the face of it, appears to be significantly higher than that for independent natural extension.

For maximal coherent sets of desirable gambles $\mathcal{M}_n \in \mathbb{M}(\mathcal{X}_n)$, $n \in N$ the strong product and the independent natural extension coincide: $\boxtimes_{n \in N} \mathcal{M}_n = \otimes_{n \in N} \mathcal{M}_n$, as clearly $m(\mathcal{M}_n) = \{\mathcal{M}_n\}$. Taking into account Proposition 23, we deduce that the strong product of maximal coherent sets of desirable gambles is not necessarily maximal; Example A.2 in the Appendix shows that there are other independent products that may strictly include the strong product.

The marginalisation properties of the strong product follow directly from those of the independent natural extension.

**Proposition 25 (Marginalisation).** *Consider coherent sets of desirable gambles $\mathcal{D}_n$ for all $n \in N$. Let $R$ be any subset of $N$, then $\mathrm{marg}_R(\boxtimes_{n \in N} \mathcal{D}_n) = \boxtimes_{r \in R} \mathcal{D}_r$.*

*Proof.* Consider any $f \in \mathcal{G}(\mathcal{X}_R)$ and observe the following chain of equivalences:

$$\begin{aligned} f \in \boxtimes_{n \in N} \mathcal{D}_n &\Leftrightarrow (\forall \mathcal{M}_n \in m(\mathcal{D}_n), n \in N) f \in \otimes_{n \in N} \mathcal{M}_n \\ &\Leftrightarrow (\forall \mathcal{M}_n \in m(\mathcal{D}_n), n \in N) f \in \otimes_{r \in R} \mathcal{M}_r \\ &\Leftrightarrow (\forall \mathcal{M}_r \in m(\mathcal{D}_r), r \in R) f \in \otimes_{r \in R} \mathcal{M}_r \\ &\Leftrightarrow f \in \boxtimes_{r \in R} \mathcal{D}_r, \end{aligned}$$

where the second equivalence follows from Proposition 17. □

Next, we show that the strong product of some coherent marginal sets of desirable gambles $\mathcal{D}_n$ is an independent product of these marginals. In order to do so, we first establish the following simple yet powerful result:

**Proposition 26.** *Let $\mathcal{D}_N^j$, $j \in J$ be any non-empty family of independent coherent sets of desirable gambles on $\mathcal{X}_N$. Then their intersection $\mathcal{D}_N := \bigcap_{j \in J} \mathcal{D}_N^j$ is an independent coherent set of desirable gambles on $\mathcal{X}_N$.*

*Proof.* Consider any disjoint subsets $I$ and $O$ of $N$, and any $x_I \in \mathcal{X}_I$. Then

$$\begin{aligned} h \in \mathrm{marg}_O(\mathcal{D}_N \rfloor x_I) &\Leftrightarrow (\forall j \in J) h \in \mathrm{marg}_O(\mathcal{D}_N^j \rfloor x_I) \\ &\Leftrightarrow (\forall j \in J) h \in \mathrm{marg}_O(\mathcal{D}_N^j) \\ &\Leftrightarrow h \in \mathrm{marg}_O(\mathcal{D}_N). \end{aligned}$$

□

**Proposition 27.** *Consider coherent marginal sets of desirable gambles $\mathcal{D}_n$ for all $n \in N$. Then their strong product $\boxtimes_{n \in N} \mathcal{D}_n$ is an independent product of these marginals.*

*Proof.* Taking into account Proposition 26, all we need to show is that the sets $\mathcal{D}_n$ are the marginals of the strong product $\boxtimes_{n \in N} \mathcal{D}_n$. This is an immediate consequence of Proposition 25. □

The strong product may strictly include the independent natural extension, as we can see from the example in Section A.3. It is an open question whether, like the independent natural extension, the strong product is associative. Although we have not been able to prove associativity in general, it is not difficult to show that it suffices to establish it for maximal sets of desirable gambles, and that one of the inclusions, namely $\boxtimes_{n \in N_1 \cup N_2} \mathcal{M}_n \subseteq (\boxtimes_{n \in N_1} \mathcal{M}_n) \boxtimes (\boxtimes_{n \in N_2} \mathcal{M}_n)$ holds because the strong product always includes the independent natural extension. We suspect, but have not been able to prove, that the converse inclusion also holds, and that the strong product is associative, taking into account that in its definition we are taking the intersection of sets of gambles determined by an associative operator (the independent natural extension).

To conclude this section, we establish a connection between the strong product of sets of desirable gambles and the eponymous notion for coherent lower previsions, studied for instance by De Cooman et al. (2011) (see also Cozman, 2012 for comments on the corresponding notion in terms of credal sets, which is sometimes called the *strong extension*). Given coherent lower previsions $\underline{P}_n$ on $\mathcal{G}(\mathcal{X}_n)$, $n \in N$, their *strong product* is the coherent lower prevision defined by

$$\underline{S}_N(f) := \inf \left\{ \times_{n \in N} P_n(f) \colon (\forall n \in N) P_n \in \mathcal{M}(\underline{P}_n) \right\}$$

for all gambles $f$ on $\mathcal{X}_N$; the intuition behind this notion, taking into account the correspondence between coherent lower previsions and sets of desirable gambles discussed in Section 2.6, is that the intersection of a family of sets of desirable gambles is closely related to taking the lower envelope of the associated family of coherent lower previsions.

If we start from linear previsions $P_n$ on $\mathcal{G}(\mathcal{X}_n)$, their strong product corresponds to their linear product $\times_{n \in N} P_n$, and it coincides also with their independent natural extension $E_N$. If we begin with coherent lower previsions $\underline{P}_n$ on $\mathcal{G}(\mathcal{X}_n)$, their strong product $\underline{S}_N$ is the lower envelope of the set of strong products determined by the dominating linear previsions.

**Theorem 28.** *Let $\mathcal{D}_n$ be coherent sets of desirable gambles in $\mathcal{G}(\mathcal{X}_n)$ for all $n \in N$, and let $\boxtimes_{n \in N} \mathcal{D}_n$ be their strong product. Consider the coherent lower previsions $\underline{P}_n$ on $\mathcal{G}(\mathcal{X}_n)$ given by $\underline{P}_n(f) := \sup\{\mu \in \mathbb{R} \colon f - \mu \in \mathcal{D}_n\}$. Then the strong product $\underline{S}_N$ of the marginal lower previsions $\underline{P}_n$, $n \in N$ satisfies*

$$\underline{S}_N(f) = \sup\left\{\mu \in \mathbb{R} \colon f - \mu \in \boxtimes_{n \in N} \mathcal{D}_n\right\}.$$

*Proof.* Assume first of all that $\mathcal{D}_n$ is a maximal coherent set of desirable gambles for all $n$ in $N$. Then it follows that $\underline{P}_n$ is a linear prevision, which we denote by $P_n$, for all $n \in N$. The strong product of the linear previsions $P_n$, $n \in N$ coincides with their linear independent product $\times_{n \in N} P_n$, which is also their independent natural extension (use Proposition 10 from De Cooman et al., 2011). Since we have proved in Theorem 21 that this is the coherent lower prevision associated with $\otimes_{n \in N} \mathcal{D}_n = \boxtimes_{n \in N} \mathcal{D}_n$, we conclude that the strong product $\boxtimes_{n \in N} \mathcal{D}_n$ is associated with the strong product of the linear previsions $P_n$.

We move next to the general case. Fix any gamble $f$ on $\mathcal{X}_N$. Consider any real number $\mu < \underline{S}_N(f)$. For any $n \in N$, consider any maximal coherent set of desirable gambles $\mathcal{M}_n \in m(\mathcal{D}_n)$, and the associated linear prevision $P_n$, then clearly $P_n \in \mathcal{M}(\underline{P}_n)$. Hence $\times_{n \in N} P_n(f) \geq \underline{S}_N(f) > \mu$, and we infer from the arguments above that then necessarily $f - \mu \in \otimes_{n \in N} \mathcal{M}_n$. Hence $f - \mu \in \boxtimes_{n \in N} \mathcal{D}_n$. This leads to the conclusion that $\underline{S}_N(f) \leq \sup \{\mu \in \mathbb{R} \colon f - \mu \in \boxtimes_{n \in N} \mathcal{D}_n\}$.

Conversely, consider any real number $\mu$ such that $f - \mu \in \boxtimes_{n \in N} \mathcal{D}_n$. Consider arbitrary $P_n \in \mathcal{M}(\underline{P}_n)$, $n \in N$, then there are maximal coherent sets of desirable gambles $\mathcal{M}_n \in m(\mathcal{D}_n)$ inducing them: let $\underline{\mathcal{D}}_n$ be the set of strictly desirable gambles that induces $P_n$, given by Eq. (3). This set is coherent by Walley (1991, Thm. 3.8.1). Consider the set $\underline{\mathcal{D}}_n \cup \mathcal{D}_n$, and let us show that it is coherent. Condition D2 holds trivially because it is satisfied by $\mathcal{D}_n$. To see that D3 holds, taking into account that both $\underline{\mathcal{D}}_n$ and $\mathcal{D}_n$ are coherent sets of gambles, and in particular cones, it suffices to show that for any gamble $f \in \underline{\mathcal{D}}_n$ and any $g \in \mathcal{D}_n$, their sum $f + g$ belongs to $\underline{\mathcal{D}}_n \cup \mathcal{D}_n$. Consider thus such gambles $f, g$. If $f \in \mathcal{G}(\mathcal{X}_n)_{>0}$, then it also belongs to $\mathcal{D}_n$ and as a consequence $f + g \in \mathcal{D}_n$; on the other hand, if $f \in \underline{\mathcal{D}}_n \setminus \mathcal{G}(\mathcal{X}_n)_{>0}$, it follows that $P_n(f) > 0$, whence $P_n(f + g) = P_n(f) + P_n(g) \geq P_n(f) + \underline{P}_n(g) > 0$, and therefore $f + g \in \underline{\mathcal{D}}_n$. Since both $\underline{\mathcal{D}}_n$ and $\mathcal{D}_n$ are coherent, we deduce from this that condition D1 also holds, and as a consequence the set $\underline{\mathcal{D}}_n \cup \mathcal{D}_n$ is indeed coherent.

Now, Theorem 5 implies that there is some maximal coherent set of desirable gambles $\mathcal{M}_n \in m(\underline{\mathcal{D}}_n \cup \mathcal{D}_n) \subseteq m(\mathcal{D}_n)$, and from Walley (1991, Thm. 3.8.3) we deduce that $\underline{\mathcal{D}}_n$ and $\mathcal{M}_n$ induce the same $P_n$ by means of Eq. (2). But then $f - \mu \in \otimes_{n \in N} \mathcal{M}_n$, and therefore $\times_{n \in N} P_n(f) \geq \mu$, using the argumentation above. Hence $\underline{S}_N(f) \geq \mu$, and therefore $\underline{S}_N(f) \geq \sup \{\mu \in \mathbb{R} \colon f - \mu \in \boxtimes_{n \in N} \mathcal{D}_n\}$. □

Together with Theorem 21 and the fact that the strong product of lower previsions may strictly dominate their independent natural extension (see Example 9.3.4 in Walley, 1991), this also shows that the strong product of marginal sets of desirable gambles may strictly include their independent natural extension. An explicit example will be given in Appendix A.3.

## 9. Conditional Irrelevance and Independence

The final step we take in this paper, consists in extending our results from irrelevance and independence to a simple but common form of conditional irrelevance and independence. Next to the variables $X_N$ in $\mathcal{X}_N$, we now also consider another variable $Y$ assuming values in a finite set $\mathcal{Y}$.

Consider two disjoint subsets $I$ and $O$ of $N$. We say that $X_I$ is *epistemically irrelevant* to $X_O$ when, conditional on $Y$, learning the value of $X_I$ does not influence or change our beliefs about $X_O$.

When does a set $\mathcal{D}$ of desirable gambles on $\mathcal{X}_N \times \mathcal{Y}$ capture this type of conditional epistemic irrelevance? Clearly, we should require that:

$$\mathrm{marg}_O(\mathcal{D} \rfloor x_I, y) = \mathrm{marg}_O(\mathcal{D} \rfloor y) \text{ for all } x_I \in \mathcal{X}_I \text{ and all } y \in \mathcal{Y}.$$

As before, for technical reasons we also allow $I$ and $O$ to be empty. It is clear from the definition above that the 'variable' $X_\emptyset$, about whose constant value we are certain, is conditionally epistemically irrelevant to any variable $X_O$. Similarly, we see that any variable $X_I$ is conditionally epistemically irrelevant to the 'variable' $X_\emptyset$. This seems to be in accordance with intuition.

Also, if $\mathcal{Y}$ is a singleton, then there is no uncertainty about $Y$ and conditioning on $Y$ amounts to not conditioning at all: epistemic irrelevance can be seen as a special case of conditional epistemic irrelevance.

We now want to argue that, conversely, there is a very specific and definite way in which conditional epistemic irrelevance statements can be reduced to simple epistemic irrelevance statements. The crucial results that allow us to establish this, are the following conceptually very simple theorem and its corollary.

**Theorem 29 (Sequential updating).** *Consider any subset $R$ of $N$, and any coherent set $\mathcal{D}$ of desirable gambles on $\mathcal{X}_N \times \mathcal{Y}$. Then*

$$(\mathcal{D}\rfloor y)\rfloor x_R = (\mathcal{D}\rfloor x_R)\rfloor y = \mathcal{D}\rfloor x_R, y \quad \text{for all } x_R \in \mathcal{X}_R \text{ and } y \in \mathcal{Y}. \tag{20}$$

*Proof.* Fix any $x_R$ in $\mathcal{X}_R$ and any $y \in \mathcal{Y}$. Clearly, all three sets in Eq. (20) are subsets of $\mathcal{G}(\mathcal{X}_{N\setminus R})$. So take any gamble $f$ on $\mathcal{X}_{N\setminus R}$, and consider the following chains of equivalences:

$$\mathbb{I}_{\{y\}}\mathbb{I}_{\{x_R\}}f \in \mathcal{D} \Leftrightarrow \mathbb{I}_{\{x_R\}}f \in \mathcal{D}\rfloor y \Leftrightarrow f \in (\mathcal{D}\rfloor y)\rfloor x_R$$
$$\mathbb{I}_{\{y\}}\mathbb{I}_{\{x_R\}}f \in \mathcal{D} \Leftrightarrow \mathbb{I}_{\{y\}}f \in \mathcal{D}\rfloor x_R \Leftrightarrow f \in (\mathcal{D}\rfloor x_R)\rfloor y$$
$$\mathbb{I}_{\{y\}}\mathbb{I}_{\{x_R\}}f \in \mathcal{D} \Leftrightarrow f \in \mathcal{D}\rfloor x_R, y.$$
$\square$

**Corollary 30 (Reduction).** *Consider any disjoint subsets $I$ and $O$ of $N$, and any coherent set $\mathcal{D}$ of desirable gambles on $\mathcal{X}_N \times \mathcal{Y}$. Then the following statements are equivalent:*

(i) $\mathrm{marg}_O(\mathcal{D}\rfloor x_I, y) = \mathrm{marg}_O(\mathcal{D}\rfloor y)$ *for all $x_I \in \mathcal{X}_I$ and all $y \in \mathcal{Y}$;*

(ii) $\mathrm{marg}_O((\mathcal{D}\rfloor y)\rfloor x_I) = \mathrm{marg}_O(\mathcal{D}\rfloor y)$ *for all $x_I \in \mathcal{X}_I$ and $y \in \mathcal{Y}$.*

This tells us that *a model $\mathcal{D}$ about $(X_N, Y)$ represents epistemic irrelevance of $X_I$ to $X_O$, conditional on $Y$ if and only if for each possible value $y \in \mathcal{Y}$ of $Y$, the model $\mathcal{D}\rfloor y$ about $X_N$ represents epistemic irrelevance of $X_I$ to $X_O$.*

Now suppose we have marginal conditional models $\mathcal{D}_n\rfloor Y$ on $\mathcal{X}_n$, $n \in N$. The notation $\mathcal{D}_n\rfloor Y$ is a concise way of representing the family of conditional models $\mathcal{D}_n\rfloor y$, $y \in \mathcal{Y}$. Then if we combine Corollary 30 and Theorem 19, we obtain the following:

**Corollary 31.** *The smallest conditionally independent product $\mathcal{D}\rfloor Y$ of the marginal models $\mathcal{D}_n\rfloor Y$ is given by $\otimes_{n \in N}(\mathcal{D}_n\rfloor Y)$, meaning that for each $y \in \mathcal{Y}$, $\mathcal{D}\rfloor y = \otimes_{n \in N}(\mathcal{D}_n\rfloor y)$.*

This also shows that calculating the conditionally independent natural extension has, in comparison with independent natural extension, an additional factor in the computational complexity that its simply linear in the number of possible values for the conditioning variable $Y$.

## 10. Conclusions

Sets of desirable gambles are more informative than coherent lower previsions, as we have shown in Section 2.6, and they are helpful in avoiding problems involving zero probabilities. Moreover, they have a simple axiomatic definition, as we have seen in Section 2.1. They have been overlooked for much of the development of the theory of imprecise probabilities, and it is only in the last five or six years that more effort is being devoted to bringing this simplifying and unifying notion to the fore.

Working with sets of desirable gambles allows us to show that the computational complexity of checking whether a gamble belongs to the independent natural extension compares favourably to that of computing the strong product, which has a complexity that is exponential in the number of variables.

Our results here also show that we can model assessments of epistemic independence easily using sets of desirable gambles, and that we can derive from them existing results for lower previsions.

Moreover, the results in Section 7 indicate that constructing global joint models (i.e. coherent sets of desirable gambles) from local ones is something that can be easily and efficiently done for the some types of credal networks (Cozman, 2000, 2005). The interpretation of the graphical structure in credal networks is usually taken to be the following: *for any node (variable), conditional on its parents, the non-parent non-descendants are strongly independent of it* (Cozman, 2000, 2005). But we can replace the assumption of strong independence with the weaker one of epistemic irrelevance, as in the work by De Cooman et al. (2010), and this tends to produce more conservative independent products.[9]

If we consider a credal network made up of $n$ unconnected nodes $X_1, \ldots, X_n$, their interpretation is then very simple: for any variable $X_k$, the remaining variables $X_1, \ldots, X_{k-1}$, $X_{k+1}, \ldots X_n$ are epistemically irrelevant to it. The expression (18) for the independent natural extension $\otimes_{k=1}^n \mathcal{D}_k$, and the reasoning behind it in Section 7, show that $\otimes_{k=1}^n \mathcal{D}_k$ is the smallest (most conservative) coherent joint set of desirable gambles that expresses the epistemic irrelevancies in the graph.

Interestingly, we can make the network slightly more complicated by looking at the developments in Section 9, which tell us that the conditionally independent natural extension $\otimes_{k=1}^n X_k \rfloor Y$ is the most conservative (conditional) joint model that reflects the independence conditions embedded in the following graphical structure:

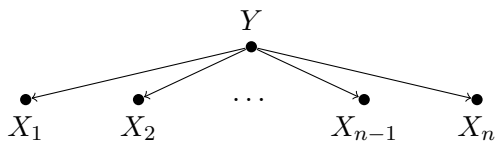

for any variable $X_k$, the remaining variables $X_1, \ldots, X_{k-1}, X_{k+1}, \ldots X_n$ are epistemically irrelevant to $X_k$, conditional on $Y$.

Now, any tree can be built up recursively using simple networks like the one above as building blocks: similarly to what is done by De Cooman et al. (2010, Section 4), we can use recursion from the leaves to the root, so that at any step we have a conditional model that we put together into a joint one using the epistemic irrelevance/independence assessments and the marginal extension theorem (Miranda & De Cooman, 2007), that allows us to combine hierarchical information. This suggests that the developments in this paper can be used to good advantage in finding efficient algorithms for inference in credal *trees* under epistemic irrelevance, using sets of desirable gambles as uncertainty models. This approach could build on the ideas proposed by De Cooman et al. (2010) and Destercke and De Cooman (2008) in the context of credal trees with lower previsions as local uncertainty models, but make them more general and also more directly amenable to simple assessment and elicitation for the local models. This would have interesting applications in dealing with hidden Markov models with imprecise transition and emission models, which are, of course, special credal trees.

---

9. See also Section 8 for more details about strong independence; we surmise that the computational complexity of dealing with strong products is worse than that for computing the independent natural extension.

We expect that generalising those algorithms towards more general credal networks (polytrees, ...) will be more difficult, and will have to rely heavily on the pioneering work of Moral (2005) on graphoid properties for epistemic irrelevance. In this sense, it would be interesting to model other assumptions of independence between variables using sets of desirable gambles, for instance intermediate assumptions between epistemic irrelevance and independence (that is, epistemic irrelevance for some pairs of sets of variables only). Moreover, algorithms for computing the irrelevant and the independent natural extension, as well as the strong product, need to be devised.

Other open problems would be to generalise our work to infinite sequences of random variables, which would allow us to deal with unbounded trees, and, as we have already discussed in the paper, to establish the associativity of the strong product and to extend our results to variables taking values on infinite spaces.

## Acknowledgments

This work was supported by SBO project 060043 of the IWT-Vlaanderen, and by project MTM2010-17844. We would like to thank the reviewers for their helpful comments.

## Appendix A. Examples

In this appendix, we have gathered a number of examples and counterexamples.

### A.1 Independent Natural Extension Need Not Preserve Maximality

Let $\mathcal{X} = \{0,1\}$ and let $\mathcal{M}$ be the subset of $\mathcal{G}(\mathcal{X})$ given by

$$\mathcal{M} := \{f \in \mathcal{G}(\mathcal{X}) \colon f(0) + f(1) > 0 \text{ or } f(0) = -f(1) > 0\}.$$

Then it is easy to see that $\mathcal{M}$ is a coherent set of desirable gambles. It is moreover maximal: if some non-zero $f \notin \mathcal{M}$, then either $f(0) + f(1) < 0$, whence $-f(0) - f(1) > 0$ and then $-f > 0$, or $f(0) = -f(1) < 0$ and then $-f(0) = f(1) > 0$, which means that $-f \in \mathcal{M}$.

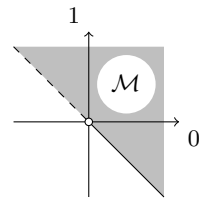

Let $N = \{1,2\}$, $\mathcal{X}_1 = \mathcal{X}_2 = \mathcal{X}$ and $\mathcal{M}_1 = \mathcal{M}_2 = \mathcal{M}$. The independent natural extension of $\mathcal{M}_1$ and $\mathcal{M}_2$ is given by

$$\mathcal{M}_1 \otimes \mathcal{M}_2 := \mathrm{posi}\left(\mathcal{G}(\mathcal{X}_{\{1,2\}})_{>0} \cup \mathcal{A}^{\mathrm{irr}}_{\{1\}\to\{2\}} \cup \mathcal{A}^{\mathrm{irr}}_{\{2\}\to\{1\}}\right)$$

$$= \left\{h_1 + h_2 \colon h_1 \in \mathcal{A}^{\mathrm{irr}}_{\{1\}\to\{2\}} \cup \{0\}, h_2 \in \mathcal{A}^{\mathrm{irr}}_{\{2\}\to\{1\}} \cup \{0\}\right\} \setminus \{0\},$$

taking into account that all non-negative gambles belong to both $\mathcal{A}^{\text{irr}}_{\{1\}\to\{2\}}$ and $\mathcal{A}^{\text{irr}}_{\{2\}\to\{1\}}$ and that $\mathcal{A}^{\text{irr}}_{\{1\}\to\{2\}}\cup\{0\}$ and $\mathcal{A}^{\text{irr}}_{\{2\}\to\{1\}}\cup\{0\}$ are convex cones. Recall that $h_1 \in \mathcal{A}^{\text{irr}}_{\{1\}\to\{2\}}\cup\{0\}$ iff $h_1(0,\cdot) \in \mathcal{M}\cup\{0\}$ and $h_1(1,\cdot) \in \mathcal{M}\cup\{0\}$, and similarly that $h_2 \in \mathcal{A}^{\text{irr}}_{\{2\}\to\{1\}}\cup\{0\}$ iff $h_2(\cdot,0) \in \mathcal{M}\cup\{0\}$ and $h_2(\cdot,1) \in \mathcal{M}\cup\{0\}$. This means that any gamble $h$ in $\mathcal{M}_1 \otimes \mathcal{M}_2$ can be expressed as

$$h(0,0) = \alpha + \epsilon, \quad h(0,1) = \beta + \mu, \quad h(1,0) = \gamma + \lambda, \quad h(1,1) = \delta + \eta,$$

where $\alpha, \ldots, \eta$ are real numbers satisfying the following constraints:

$$\alpha + \beta > 0 \text{ or } \alpha = -\beta \geq 0$$
$$\gamma + \delta > 0 \text{ or } \gamma = -\delta \geq 0$$
$$\epsilon + \lambda > 0 \text{ or } \epsilon = -\lambda \geq 0$$
$$\mu + \eta > 0 \text{ or } \mu = -\eta \geq 0$$
$$\max\{\alpha, \gamma, \epsilon, \mu\} > 0.$$

Then the gamble $h$ given by $h(0,0) = h(1,1) = -1$ and $h(0,1) = h(1,0) = 1$ does not belong to $\mathcal{M}_1 \otimes \mathcal{M}_2$: since $h(0,0) + h(0,1) + h(1,0) + h(1,1) = 0$, we should have $\alpha = -\beta \geq 0$, $\gamma = -\delta \geq 0$, $\epsilon = -\lambda \geq 0$ and $\mu = -\eta \geq 0$, and this implies that $h(0,0) \geq 0$, a contradiction. But $-h$ does not belong to $\mathcal{M}_1 \otimes \mathcal{M}_2$ either, because $h(0,0) + h(0,1) + h(1,0) + h(1,1) = 0$ similarly implies that $-h(1,1) \leq 0$. Hence, the independent natural extension of $\mathcal{M}_1$ and $\mathcal{M}_2$ is not maximal.

### A.2 A Maximal Independent Product of Maximal Sets

Next, we construct an example of an independent product of maximal sets that is again maximal.

Consider the spaces $\mathcal{X}_1$ and $\mathcal{X}_2$, and the maximal marginal coherent sets of desirable gambles $\mathcal{M}_1$ and $\mathcal{M}_2$ as in Section A.1. Now consider the set of desirable gambles $\mathcal{M}'$ defined by

$$\mathcal{M}' := \{h \in \mathcal{G}(\mathcal{X}_N): h(0,0) + h(0,1) + h(1,0) + h(1,1) > 0\}$$
$$\cup \{h \in \mathcal{G}(\mathcal{X}_N): h(0,0) + h(0,1) + h(1,0) + h(1,1) = 0 \text{ and }$$
$$[h(0,0) > 0 \text{ or } h(0,0) = 0, h(0,1) > 0 \text{ or } h(0,0) = h(0,1) = 0, h(1,0) > 0]\}.$$

We first show that $\mathcal{M}_1 \otimes \mathcal{M}_2 \subseteq \mathcal{M}'$. According to the discussion in Section A.1, any gamble $h$ in $\mathcal{M}_1 \otimes \mathcal{M}_2$ satisfies $h(0,0) = \alpha + \epsilon$, $h(0,1) = \beta + \mu$, $h(1,0) = \gamma + \lambda$, and $h(1,1) = \delta + \eta$, where in particular

$$\min\{\alpha + \beta, \gamma + \delta, \epsilon + \lambda, \mu + \eta\} \geq 0,$$

whence

$$h(0,0) + h(0,1) + h(1,0) + h(1,1) = (\alpha + \epsilon) + (\beta + \mu) + (\gamma + \lambda) + (\mu + \eta)$$
$$= (\alpha + \beta) + (\gamma + \delta) + (\epsilon + \lambda) + (\mu + \eta) \geq 0.$$

If $h(0,0) + h(0,1) + h(1,0) + h(1,1) = 0$, this implies that $\alpha + \beta = \gamma + \delta = \epsilon + \lambda = \mu + \eta = 0$, and therefore, again looking at the characterisation of $\mathcal{M}_1 \otimes \mathcal{M}_2$ in Section A.1, that

$\alpha = -\beta \geq 0$, $\gamma = -\delta \geq 0$, $\epsilon = -\lambda \geq 0$ and $\mu = -\eta \geq 0$ This implies in particular that $h(0,0) = \alpha + \epsilon \geq 0$. So we see that either $h(0,0) > 0$, and in that case $h \in \mathcal{M}'$, or $h(0,0) = 0$. But this implies that $\alpha = \epsilon = \beta = \lambda = 0$. And then $h(0,1) = \mu \geq 0$, so again we have either $h(0,1) > 0$, in which case $h \in \mathcal{M}'$, or $h(0,1) = \mu = \delta = 0$. But now it follows from the conditions imposed on the $\alpha,\ldots,\eta$ in Section A.1 that $h(1,0) = \gamma > 0$, which again means that $h$ belongs to $\mathcal{M}'$. So, indeed, $\mathcal{M}_1 \otimes \mathcal{M}_2 \subseteq \mathcal{M}'$.

We now show that $\mathcal{M}'$ is a maximal coherent set of desirable gambles. It is easy to see that $\mathcal{M}'$ is coherent. To show that it is maximal, consider any non-zero gamble $h$ in $\mathcal{G}(\mathcal{X}_{\{1,2\}})$; then there are three possibilities. If $h(0,0) + h(1,0) + h(0,1) + h(1,1) > 0$, then $h \in \mathcal{M}'$ and $-h \notin \mathcal{M}'$. If $h(0,0) + h(1,0) + h(0,1) + h(1,1) < 0$, then $-h \in \mathcal{M}'$ and $h \notin \mathcal{M}'$. And if $h(0,0) + h(1,0) + h(0,1) + h(1,1) = 0$, then exactly one of $h, -h$ belongs to $\mathcal{M}'$.

To conclude, note that $\mathcal{M}'$ is an independent product of $\mathcal{M}_1$ and $\mathcal{M}_2$ because of Proposition 24.

### A.3 The Strong Product May Strictly Include The Independent Natural Extension

The following is an adaptation of an example by Walley (1991, Example 9.3.4).

Consider $\mathcal{X} = \{0, 1\}$ and let $\underline{P}$ be the coherent lower prevision determined by $\underline{P}(\{0\}) = 2/5$ and $\underline{P}(1) = 1/2$, so we have for all $f \in \mathcal{G}(\mathcal{X})$ that:

$$\underline{P}(f) = \min\left\{\frac{1}{2}f(0) + \frac{1}{2}f(1), \frac{2}{5}f(0) + \frac{3}{5}f(1)\right\}.$$

With $\underline{P}$ we can associate the coherent set of (strictly) desirable gambles by Eq. (3):

$$\underline{\mathcal{D}} := \{f \colon f > 0 \text{ or } \underline{P}(f) > 0\}.$$

Now let $N = \{1, 2\}$, $\mathcal{X}_1 = \mathcal{X}_2 = \mathcal{X}$ and $\mathcal{D}_1 = \mathcal{D}_2 = \underline{\mathcal{D}}$. Consider the gamble $h$ on $\mathcal{X}_{\{1,2\}}$ determined by

$$h(0,0) = h(1,1) = \frac{51}{100}, \quad h(0,1) = h(1,0) = -\frac{49}{100}.$$

To see that $\mathcal{D}_1 \otimes \mathcal{D}_2$ is strictly included in $\mathcal{D}_1 \boxtimes \mathcal{D}_2$, we will show that $h$ belongs to $\mathcal{D}_1 \boxtimes \mathcal{D}_2$ but not to $\mathcal{D}_1 \otimes \mathcal{D}_2$.

For the latter claim, consider any gambles $h_1 \in \mathcal{A}^{\mathrm{irr}}_{\{1\}\to\{2\}}$ and $h_2 \in \mathcal{A}^{\mathrm{irr}}_{\{2\}\to\{1\}}$, and assume ex absurdo that $h \geq h_1 + h_2$. Then we see that $(h_1 + h_2)(0,0) = \alpha + \epsilon$, $(h_1 + h_2)(0,1) = \beta + \mu$, $(h_1 + h_2)(1,0) = \gamma + \lambda$ and $(h_1 + h_2)(1,1) = \delta + \eta$, where the real numbers $\alpha, \ldots, \eta$ must satisfy the following constraints:

$$\max\{\alpha, \beta\} > 0 \text{ and } \min\left\{\frac{1}{2}\alpha + \frac{1}{2}\beta, \frac{2}{5}\alpha + \frac{3}{5}\beta\right\} \geq 0$$

$$\max\{\gamma, \delta\} > 0 \text{ and } \min\left\{\frac{1}{2}\gamma + \frac{1}{2}\delta, \frac{2}{5}\gamma + \frac{3}{5}\delta\right\} \geq 0$$

$$\max\{\epsilon, \lambda\} > 0 \text{ and } \min\left\{\frac{1}{2}\epsilon + \frac{1}{2}\lambda, \frac{2}{5}\epsilon + \frac{3}{5}\lambda\right\} \geq 0$$

$$\max\{\mu, \eta\} > 0 \text{ and } \min\left\{\frac{1}{2}\mu + \frac{1}{2}\eta, \frac{2}{5}\mu + \frac{3}{5}\eta\right\} \geq 0.$$

As a consequence,

$$\frac{2}{5}(\alpha + \epsilon) + \frac{3}{5}(\beta + \gamma + \delta + \mu + \lambda + \eta)$$
$$= (\frac{2}{5}\alpha + \frac{3}{5}\beta) + (\frac{2}{5}\epsilon + \frac{3}{5}\lambda) + \frac{6}{5}(\frac{1}{2}\gamma + \frac{1}{2}\delta) + \frac{6}{5}(\frac{1}{2}\mu + \frac{1}{2}\eta) \geq 0,$$

but on the other hand

$$\frac{2}{5}(\alpha + \epsilon) + \frac{3}{5}(\beta + \gamma + \delta + \mu + \lambda + \eta)$$
$$\leq \frac{2}{5}h(0,0) + \frac{3}{5}(h(0,1) + h(1,0) + h(1,1)) = -\frac{39}{500},$$

a contradiction. This implies that $h$ does not belong to $\mathcal{D}_1 \otimes \mathcal{D}_2$.

For the former claim, consider arbitrary maximal coherent set of desirable gambles $\mathcal{M}_1 \in m(\mathcal{D}_1)$ and $\mathcal{M}_2 \in m(\mathcal{D}_2)$. Then it follows from the discussion in Section 2.6 that $\mathcal{M}_1$ induces a linear prevision $P_1 \geq \underline{P}_1$, and that $\mathcal{M}_2$ induces a linear prevision $P_2 \geq \underline{P}_2$. But then it follows from the discussion in Example 9.3.4 by Walley (1991) that

$$(P_1 \times P_2)(h) \geq \frac{1}{100} > 0,$$

which tells us that $h$ belongs to the set of strictly desirable gambles that induces $P_1 \times P_2$, because of Eq. (3). Since that is the smallest coherent set of desirable gambles that induces $P_1 \times P_2$, and since $\mathcal{M}_1 \otimes \mathcal{M}_2$ is another such set, by Theorem 28, we deduce that $h \in \mathcal{M}_1 \otimes \mathcal{M}_2$. It follows that indeed $h \in \mathcal{D}_1 \boxtimes \mathcal{D}_2$.